%% file: main.tex
\pdfoutput=1
\documentclass[11pt]{article}

\usepackage[preprint]{acl}

\usepackage{hyperref}
\usepackage{subcaption}
\usepackage{booktabs}
\usepackage{tabularx}
\usepackage{amssymb}
\usepackage{enumitem}
\usepackage{float} 

\usepackage{amsmath}
\usepackage{bbm} % for \mathbbm{1}

\usepackage{algorithm}
\usepackage{algpseudocode}

\usepackage{xcolor}

\usepackage{times}
\usepackage{latexsym}

\usepackage[T1]{fontenc}
\usepackage[utf8]{inputenc}

\usepackage{microtype}

\usepackage{inconsolata}

\usepackage{graphicx}
\usepackage{amsmath}
\usepackage{lipsum}

\usepackage{xargs}                      % Use more than one optional parameter in a new commands
\newif\ifshowdraft

\showdrafttrue

\title{Discriminating Form and Meaning in Multilingual Models with Minimal-Pair ABX Tasks}

\author{
  Maureen de Seyssel\thanks{Equal contribution} \quad
  Jie Chi\footnotemark[1] \quad
  Skyler Seto \quad
  Maartje ter Hoeve \\
  {\bf Masha Fedzechkina \quad Natalie Schluter} \\
  Apple \\
  \texttt{\{mdeseyssel,jchi2\}@apple.com}
}

\begin{document}
\maketitle

\begin{abstract}

We introduce a set of training-free ABX-style discrimination tasks to evaluate how multilingual language models represent language identity (form) and semantic content (meaning). Inspired by speech processing, these zero-shot tasks measure whether minimal differences in representation can be reliably detected. This offers a flexible and interpretable alternative to probing. Applied to XLM-R \cite{conneau2020unsupervised} across pretraining checkpoints and layers, we find that language discrimination declines over training and becomes concentrated in lower layers, while meaning discrimination strengthens over time and stabilizes in deeper layers. We then explore probing tasks, showing some alignment between our metrics and linguistic learning performance.  
Our results position ABX tasks as a lightweight framework for analyzing the structure of multilingual representations.
\end{abstract}

% e further relate these scores to probing tasks: language discrimination correlates with POS and cross-lingual NER accuracy, while meaning discrimination shows no clear alignment with task performance.  

\section{Introduction}

Multilingual Transformer models such as mBERT
\cite{devlin2019bert} and XLM-R \cite{conneau2020unsupervised} have become essential tools for cross-lingual NLP. Trained on large concatenated corpora spanning dozens of languages, these models learn representations that support transfer across languages even in the absence of explicit cross-lingual supervision \cite[etc.]{wu2019beto, conneau2020emerging, xue2021mt5, philippy2023towards}. Despite their success, it remains unclear how these models internally organize linguistic form and shared meaning. Prior work suggests that multilingual models encode both language-specific information (e.g., surface forms, word order) and language-agnostic features (e.g., semantic content), but the nature and interaction of these representations is not fully understood. Recent analyses also identify language-specific internal mechanisms in multilingual LLMs (e.g., controllable language-specific neurons) and investigate which language representations models pivot through during processing \citep{kojima2024multilingual,zhong2024beyond}.
These encoding choices shape generalization and transfer behavior, including both positive effects (e.g., shared structure benefiting low-resource languages) and negative ones, such as the \textit{curse of multilinguality} \cite{conneau2020unsupervised}, where performance degrades due to interference across languages. 

Understanding how form and meaning are represented, and how this balance evolves during pretraining, is essential to explain and improve cross-lingual transfer. If a model strongly encodes language identity, it may better avoid interference between closely related languages\footnote{In fact, it was found that forcing some sort of separation in multilingual models can help somewhat alleviate these negative interferences \cite{pfeiffer2022lifting,blevins2024breaking,xu2024x,huang2024modular}.}. Conversely, if it aligns meanings across languages, it may support more effective semantic generalization. To explore this balance, we ask: \textit{How are languages represented at the form level? How well do models encode shared meanings? And how do these properties evolve across training?}

Previous work has often relied on probing tasks to investigate such questions. While useful, probing requires training classifiers on top of frozen representations, and results are sensitive to probe design and task setup \cite{belinkov2022probing,hewitt2019designing,voita2020information}. This makes it difficult to isolate what is truly encoded by the model versus what is learnable with supervision.

We propose a zero-shot alternative: ABX-style discrimination tasks that directly measure model representation structure without additional training.  Originating in speech processing \cite{schatz2013evaluating}, ABX tasks evaluate whether a model reliably discriminates minimal contrasts: given a triplet ($A$, $B$, $X$), is $X$ closer to $A$ or $B$? By designing minimal pairs that differ only in language or in meaning, we isolate and quantify how well models distinguishes these dimensions. Because they are contrastive, zero-shot, and training-free, these metrics can be applied across languages, checkpoints, and architectures with minimal adaptation.

% \vspace{0.8em}
\noindent Our contributions are as follows:
\begin{enumerate}[noitemsep]
\item We propose a training-free, ABX-style framework for analyzing multilingual representations by contrasting minimal pairs. Our tasks are designed to isolate language identity (form) and semantic content (meaning), offering a complementary alternative to traditional probing methods.

\item We apply this framework to XLM-R across 36 languages and 630 language pairs, analyzing all pretraining checkpoints and layers. We show that language and meaning discrimination evolve in parallel but are not mutually exclusive: different layers vary in the degree to which they encode each axis. 

\item We relate ABX discrimination scores to downstream performance on POS tagging, NER, and NLI. We find that form-oriented tasks correlate more strongly with language discrimination, while NLI, a semantic task, shows no consistent relationship to either axis, highlighting a disconnect between task performance and intrinsic representational structure.
\end{enumerate}

% We should note here that the goal in this paper is to introduce and validate an ABX-based framework for multilingual representation analysis, not to produce a broad multi-model benchmark. We therefore focus the empirical study on XLM-R \cite{conneau2020unsupervised}, which offers (i) publicly available pretraining checkpoints at fine granularity \cite{blevins2022analyzing}, crucial for tracking training dynamics, and (ii) an encoder architecture widely used for sentence-level probing tasks (POS, NER, NLI), enabling clean comparisons across layers. The ABX framework itself is model-agnostic and can be easily applied to decoder-only and encoder–decoder models given a consistent sentence representation. Extending the layer-level analyses to recent LLMs (e.g., decoder-only models) is a valuable next step, but is orthogonal to the present paper’s methodological contribution.

We note that the goal of this paper is to introduce and validate an ABX-based framework for multilingual representation analysis, not to produce a broad multi-model benchmark. Accordingly, we focus on XLM-R \cite{conneau2020unsupervised}, which (i) provides publicly available checkpoints at fine granularity \cite{blevins2022analyzing}, crucial for tracking training dynamics, and (ii) uses an encoder architecture widely adopted for probing tasks (POS, NER, NLI), enabling clean layer-wise comparisons. The framework itself is model-agnostic and can be applied to decoder-only or encoder–decoder models with consistent sentence representations. Extending the analysis to recent LLMs is an important next step, but beyond this paper’s methodological scope.

\section{Related Work}

Multilingual language models are expected to support cross-lingual generalization by encoding both language-specific form and shared semantic content. However, existing evaluation methods typically focus on one of these dimensions in isolation.
 This section reviews prior work on
 analyzing multilingual representations and highlights the need for a unified, training-free framework that jointly evaluates both language identity and meaning in a controlled, contrastive setting.

\subsection{Evaluating Form and Content in Multilingual Representations} 

% probing in multilingual representations?

Multilingual pretrained language models aim to map diverse languages into a shared embedding space. This allows for zero-shot and cross-lingual transfer, but raises the question of  how these models balance language-specific and language-agnostic features during training.

Content-focused evaluations typically  focus on cross-lingual alignment, using methods such as translation retrieval to measure whether semantically equivalent inputs in different languages are mapped to nearby embeddings \cite{sundar2025steering, pires2019multilingual, libovicky2020language,hu2020xtreme}. Models like LASER  \cite{artetxe2019massively} and LaBSE \cite{feng2022language} are explicitly trained to optimize such alignment. More recent work introduces contrastive alignment scores such as DALI \cite{ravisankar2025can} to better capture meaning equivalence. However, these approaches abstract away from language identity and provide little insight into how models handle surface-form distinctions across languages.

Conversely, form-focused evaluations examine how well a model encodes language identity. Clustering analyses show that multilingual embeddings often group by language or script, particularly in lower layers \cite{libovicky2020language, choenni2022investigating}.  Classifiers trained on frozen representations can often identify input language with high accuracy \cite{choenni2022investigating}, but this depends on probe training and may not reflect the geometry of the representation space itself.

These two evaluation paradigms have remained largely separate. To our knowledge, no existing method allows for simultaneous, controlled evaluation of both dimensions without relying on task-specific training. As a result, we lack a unified evaluation framework that can directly assess both dimensions under comparable, controlled conditions.

\subsection{Training Dynamics and Linguistic Emergence in Multilingual Models}
Several studies have examined how multilingual representations evolve during pretraining. \citet{blevins2022analyzing} tracked the emergence of linguistic knowledge in XLM-R \cite{conneau2020unsupervised}, showing that different properties emerge at different layers and stages, and that the best-performing checkpoint varies across languages and tasks. Other studies have shown that multilingual models sometimes internally pivot through high-resource languages like English when processing low-resource inputs \cite{wendler2024llamas, schut2025multilingual}, while other research suggests that these models juggle both language-specific and language-neutral features \cite{tang2024language,libovicky2020language,tanti2021language}.
These works highlight the complex interplay between form and content in multilingual models and how this balance shifts over time. However, they again mainly rely on task-specific probes or downstream evaluations, which do not offer a way to disentangle form and content in a direct, unsupervised way.

\subsection{Prior Uses of ABX Evaluation}
The ABX framework offers a contrastive, classifier-free means of evaluating representational structure in a controlled, unsupervised setting. 
 Originally developed in speech processing and psycholinguistics \cite{schatz2013evaluating, schatz2014evaluating,schatz2016abx}, ABX tests ask whether a test item $X$ is more similar (in embedding space) to a reference item $A$ or to an alternative $B$. By controlling the design of $A$, $B$, and $X$, ABX evaluations can isolate specific factors of interest (such as phoneme identity in speech) while holding others constant (e.g., speaker, context) \cite[etc.]{versteegh2015zero, dunbar2017zero, hallap2022evaluating, sicherman2023analysing}. ABX tasks have proven robust to variability from other categorical structures, enabling reliable measurement of the target factor even when other linguistic or speaker-related properties vary \cite{schatz2016abx}.

While ABX has primarily been applied to phoneme discrimination, recent work has begun adapting it to the other tasks, testing models' ability to discriminate between languages \cite{carbajal2016modeling, de2020does, de2023unsupervised}, speakers \cite{thorburn2019quantitative,de2022language} and to evaluate syntactic or semantic distinctions \cite{algayres2022dp, algayres2023generative}.

\vskip 1em
Our work builds on this foundation by adapting ABX discrimination to text-based multilingual models. We propose a set of zero-shot tasks that independently measure sensitivity to language identity and semantic content using minimal contrast triplets. To our knowledge, this is the first unified, training-free framework that systematically isolates and evaluates these two core dimensions of multilingual representation.

\section{Our ABX Discrimination Framework}

Understanding how multilingual models structure linguistic information in their internal representations is key to explaining their interactions between different languages, and to a further extent their generalization behaviors.
To directly assess the intrinsic structure of multilingual representations without relying on the pitfalls of extrinsic evaluation, we adapt the ABX discrimination paradigm, originally developed for evaluating speech embeddings, to the text domain.

\begin{figure}[htpb]
  \centering
  \includegraphics[width=0.55\columnwidth]{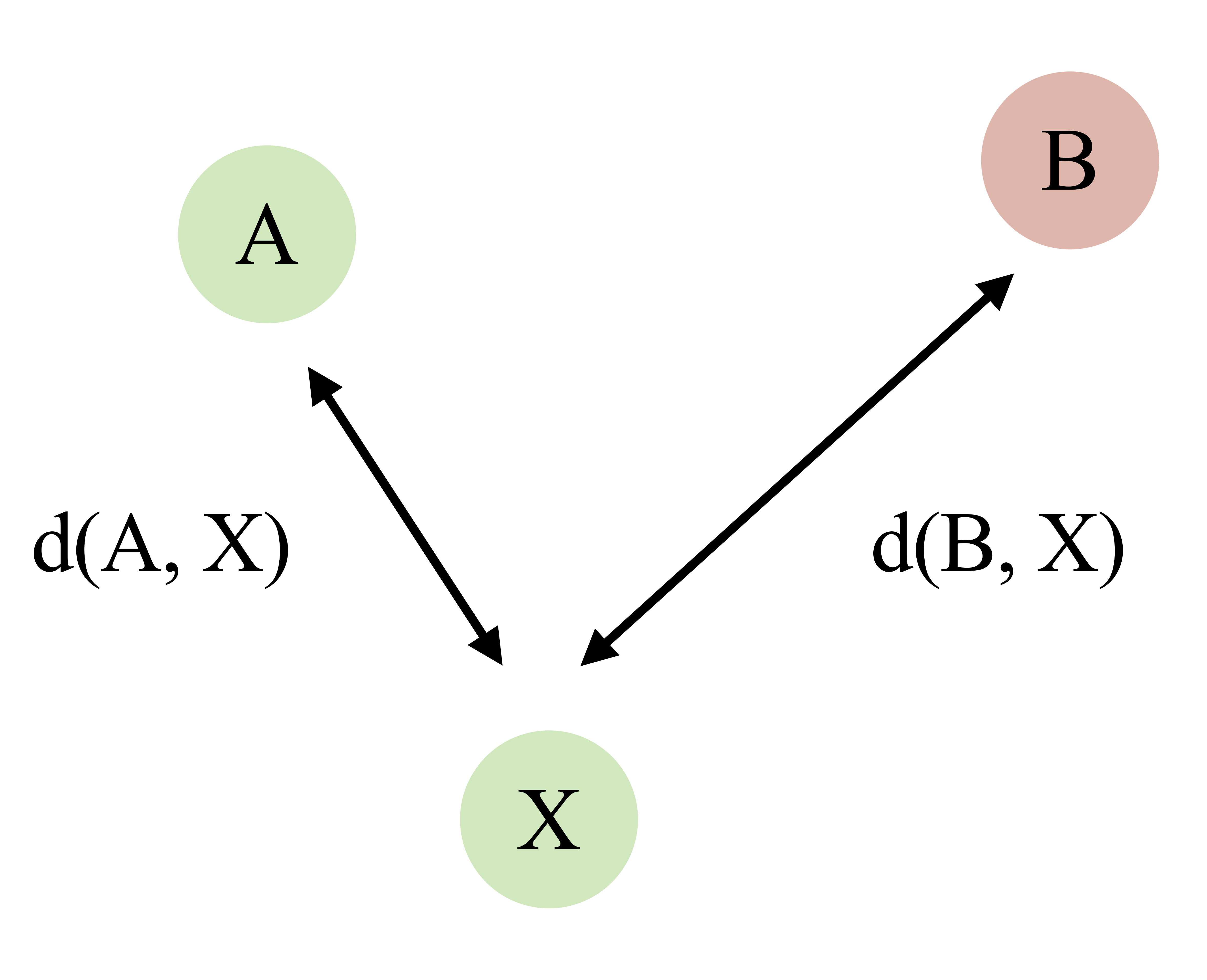}
  \caption{Illustration of the ABX discrimination task. $A$ and $X$ share the target variable, whereas $B$ differs. Control variables may be included, with $A$ and $B$ sharing the same control variable.}
  \label{fig:abx_setups}
\end{figure}

In the  original ABX framework \cite{schatz2013evaluating,schatz2014evaluating, schatz2016abx}, illustrated in Figure~\ref{fig:abx_setups}, three items ($A$, $B$, $X$) are presented, with $A$ and $B$ belonging to different categories, and $X$ matching the category of either $A$ or $B$. A model is successful when $X$ is closer (according to a distance metric $d$ in embedding space) to the item that shares its category. That is, for each triplet, a correct decision is recorded when $d(X, A) < d(X, B)$, with $X$ and $A$ sharing the same category. The score for a given triplet is computed as :

\[
\operatorname{score}(A, B, X) = \mathbbm{1}{[\, d(X, A) < d(X, B) \,]}
\]

where $\mathbbm{1}$ denotes the indicator function. The overall ABX score is the average success rate across all triplets.  Importantly, control variables can be introduced to eliminate bias from confounding factors. In that case, both $A$ and $B$ share the same control variable to ensure that the discrimination is based solely on the variable of interest.

The ABX score reflects the proportion of correct decisions, with higher scores indicating better discrimination. We apply this setup to sentence embeddings extracted from XLM-R at various layers and checkpoints, where each sentence is represented by the mean-pooled embedding of its subword tokens. Cosine similarity is used as the distance metric\footnote{We choose cosine as it is the standard metric in many embedding-based evaluations, particularly in multilingual sentence retrieval and alignment tasks \citep[e.g.,][]{ravisankar2025can,sundar2025steering,mohammadshahi2019aligning}. Cosine is well-suited to measuring relative orientation in high-dimensional spaces and is less sensitive to differences in embedding magnitude, which makes it particularly effective for comparing representations across languages and layers.}.

We propose two ABX variants for studying multilingual language models: \textit{language discrimination} (LD) and \textit{meaning discrimination} (MD).
Both tasks leverage paired multilingual data and are constructed to isolate either language identity or semantic content while controlling for the other. These tasks enable zero-shot, training-free evaluation of key representational properties in multilingual models.

We present both tasks below; see Appendix~\ref{appendix:abx-setups} for further illustrations and examples, and Appendix~\ref{appendix:pseudocode} for pseudocode.

\paragraph{Language Discrimination}
In the LD task, the objective is to assess whether the model can distinguish between embedding representations from different languages while controlling for meaning. In other words, the focus is on determining whether the form of the language is encoded in the representations sufficiently to discriminate between languages. Triplets are constructed as follows: $X$ comes from language $L_1$ and carries meaning $M_1$; $A$, the target, is also from language $L_1$ but conveys a different meaning $M_2$; and $B$, the distractor, is from another language $L_2$ but shares the same meaning $M_2$ as $A$, hence controlling for meaning. For example, $X$ might be an English sentence about the weather, $A$ another English sentence with a different meaning, and $B$ the French translation of $A$ (see Appendix~\ref{appendix:abx-setups} for a more explicit illustration). The task is considered successful if the model leads to the distance between $A$ and $X$ to be smaller than that between $B$ and $X$.

\paragraph{Meaning Discrimination}
In the MD task, we test whether the model captures differences in meaning while holding language constant. The goal is to evaluate whether semantic content is encoded in the representations independently of surface form. Triplets are constructed such that $A$ and $X$ share the same meaning $M_1$ but come from different languages ($L_1$ and $L_2$), while $B$ is in the same language as $A$ ($L_2$) but conveys a different meaning $M_2$. For instance, $X$ could be an English sentence about the weather, $A$ its French translation, and $B$ a French sentence with an unrelated meaning. The model is considered successful if it places $X$ closer to $A$ than to $B$, indicating that it encodes semantic similarity across languages, beyond surface-level language identity.

\vskip 0.3cm
While LD primarily probes the presence of language-specific information, MD offers a more direct lens on semantic similarity. High MD scores, especially across languages, suggest that the model encodes meaning in a way that is at least partially language-agnostic. As such, MD may serve as a proxy for cross-lingual semantic alignment within the representation space. In fact, we show in Section~\ref{sec:abx:validation} that a standard cross-lingual retrieval task, commonly used to assess such alignment, correlates highly with our MD task, supporting the idea that MD captures cross-lingual alignment\footnote{The high correlation does not imply they are identical. Our ABX MD task targets the same underlying ability under more controlled conditions. Instead of ranking many candidates, MD ABX uses contrastive triplets that isolate semantic differences while tightly controlling for language.}. We do not perform a similar analysis for LD, as no existing metric captures the specific abilities assessed by our language ABX task.

\section{Discrimination dynamics in a multilingual model}

\subsection{General Experimental setup}
\paragraph{Model}
To study pretraining dynamics in a multilingual setting, we use the base version of XLM-R \cite{conneau2020unsupervised} (L = 12, H = 768, A = 12, 270M parameters), a widely used multilingual masked language model. Specifically, we rely on the 39 checkpoints released by \citet{blevins2022analyzing}, who retrained XLM-R from scratch in order to examine the evolution of language representations during pretraining\footnote{Checkpoints were saved every 5k training steps up to step 50k and then every 50k steps until the final 1.5M update. Details of the pretraining scheme can be found in \citet{blevins2022analyzing}.}. All evaluations and analyses in this work are based on the representations from these checkpoints.

\paragraph{ABX Languages and Dataset}\label{sec:abx:data_and_methods}
We construct ABX triplets and perform evaluations using the WMT24++ dataset \cite{deutsch2025wmt24++}, a multilingual corpus of 55 languages with parallel sentence-level alignments across all language pairs\footnote{Here, “alignment” refers strictly to parallel corpus sentence alignment ensuring meaning equivalence.}. From this corpus, we select 36 languages spanning a broad range of families, scripts, and typological features (see Appendix~\ref{app:languages} for the complete list). This selection yields 630 unordered language pairs. Triplets are sampled randomly from aligned sentence pairs, ensuring that each triplet satisfies the relevant ABX condition (form or meaning), and that the sample size is sufficient to ensure broad and unbiased coverage. For each evaluation mode and language pair, we generate approximately 100{,}000 triplets.
Unless stated otherwise, we report discrimination scores averaged across all layers. In most analyses, we present scores separately by checkpoint to track how discrimination abilities evolve during training. In addition to language-pair scores, we compute a global LD or MD score for each language, defined as the average across all pairings with the other 35 languages. These global metrics offer a higher-level view of how well a language is discriminated or semantically aligned within the multilingual space.

\paragraph{Validation of ABX Metrics}\label{sec:abx:validation}

To validate our metrics, we perform two control analyses. First, we confirm that both ABX scores return values at or near chance (0.5) under a "baseline" setup, where the variable of interest (language for LD, meaning for MD) is held constant across all three elements of the triplet. This serves as a sanity check to rule out bias in the construction of the triplets or evaluation procedure.
Second, we compare MD scores at the final checkpoint with performance on a standard cross-lingual retrieval task. Following the setup of \citet{sundar2025steering}, we compute, for each language pair $(L_1, L_2)$, the top-1 accuracy of retrieving the most semantically similar sentence in language $L_2$ given a sentence in language $L_1$. The retrieval pool consists of $L_2$ candidates from the WMT24 dataset (and vice versa). While retrieval tasks typically rely on mean-pooled representations from the final layer, our ABX evaluations average scores across all layers. Despite this difference, we find a strong correlation between the two metrics (Pearson $r = 0.77$, $p<0.001$)\footnote{When both use last-layer embeddings, $r = 0.73$; when comparing last-layer retrieval to all-layer ABX, Pearson drops to $r = 0.53$, but remains significant ($p<0.001$).}. This supports the validity of ABX as a proxy for semantic alignment. Importantly, ABX goes further by explicitly controlling for surface form (in the case of MD), enabling a more fine-grained assessment of the model’s semantic representations.

\subsection{Experiments}

\begin{figure}[t]
  \includegraphics[width=\columnwidth]{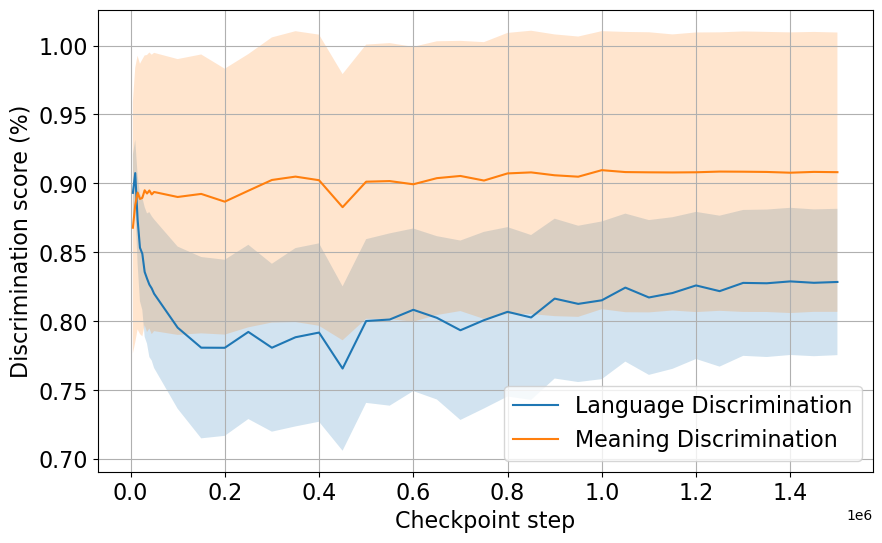}
    \caption{Language and meaning ABX discrimination scores across checkpoints (averaged over layers and all language pairs). Baseline score is 0.5.}
    
  \label{fig:ABX_across_checkpoints}
\end{figure}

\begin{figure}[t]
  \includegraphics[width=\columnwidth]{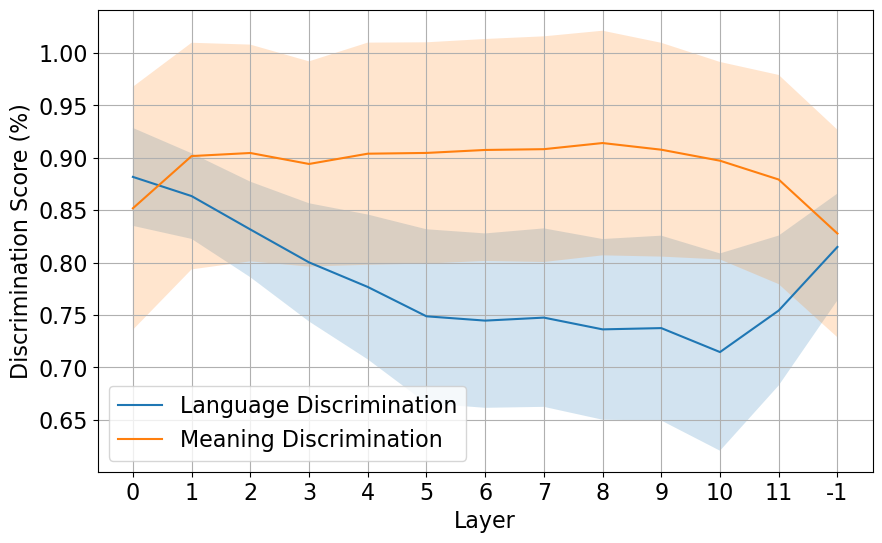}
  \caption{Language and meaning ABX discrimination scores across layers (averaged over all language pairs) for the last checkpoint (step 150,000)}
  \label{fig:ABX_across_layers}
\end{figure}

\begin{figure*}[t]
  \includegraphics[width=0.48\linewidth]{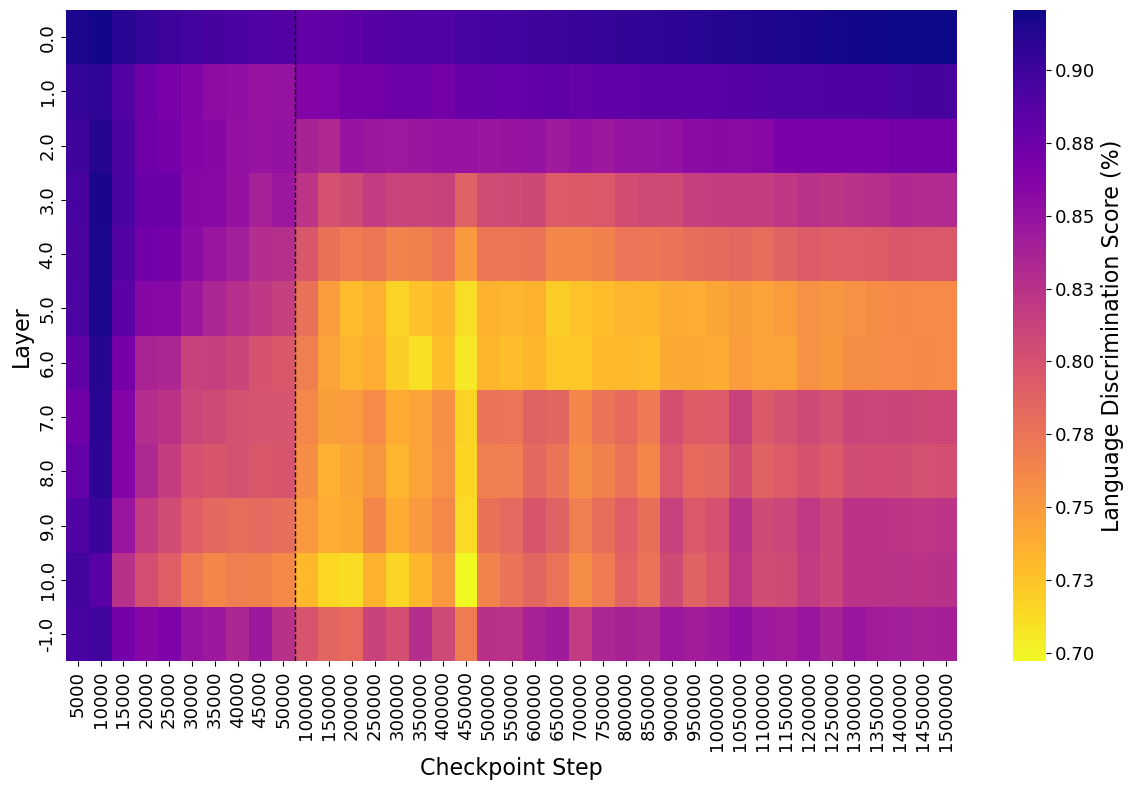} \hfill
  \includegraphics[width=0.48\linewidth]{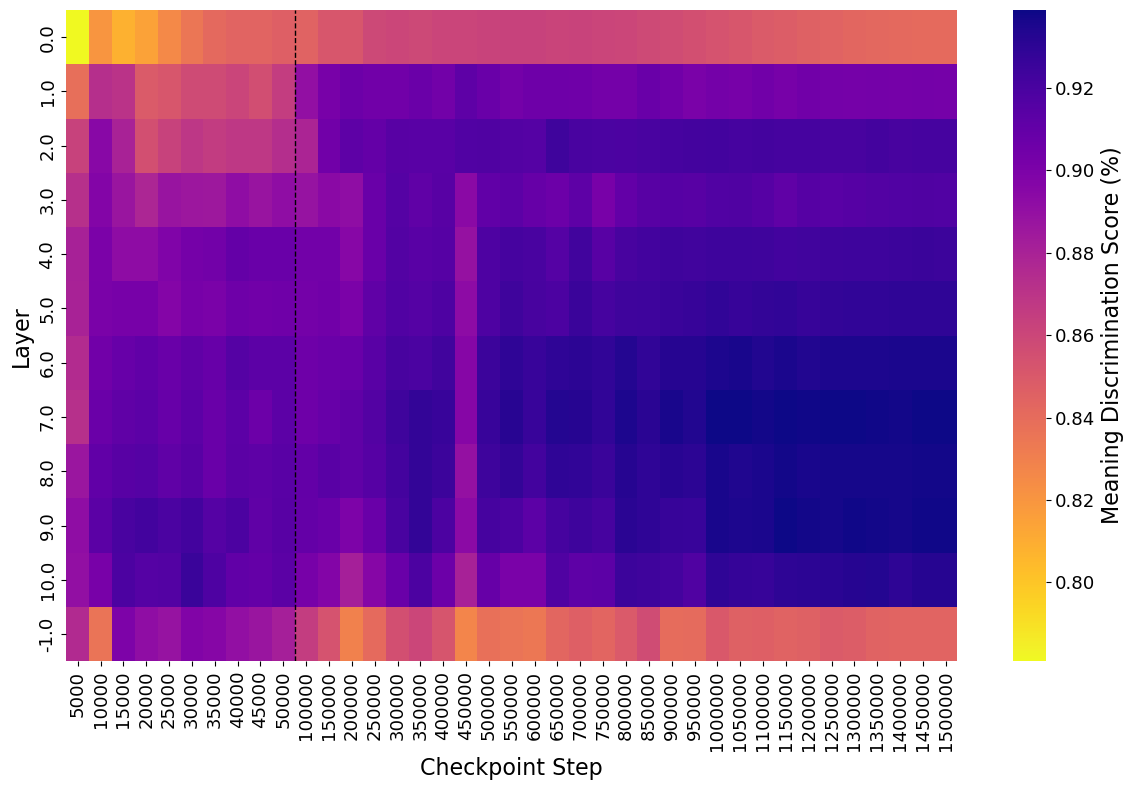}
  \caption{Evolution of ABX discrimination scores across model checkpoints and layers. Dark regions indicate higher discrimination scores. (Left: LD; right: MD)}
  \label{fig:ABX_heatmaps}
\end{figure*}

We begin by analyzing how the model’s ability to discriminate between language identity (form) and semantic meaning (content) evolves during training. Figures~\ref{fig:ABX_across_checkpoints}, \ref{fig:ABX_across_layers}, and \ref{fig:ABX_heatmaps} present complementary views of these dynamics across checkpoints and layers.

\paragraph{Checkpoint-level evolution.} Figure~\ref{fig:ABX_across_checkpoints} shows the evolution of average LD and MD scores across checkpoints, aggregated over all language pairs. First of all, we can see that all scores are consistently above the 0.5 baseline, meaning that the model, at all checkpoints, can discriminate between languages and meanings (cross-lingually) to some extent.
LD score declines rapidly during early training steps and gradually recovers in later stages, while MD score steadily improves. This suggests that as training progresses, the model increasingly prioritizes semantic abstraction over explicit language-specific cues.

We also observe a negative correlation between the two measures when considering all language pairs across checkpoints (Spearman’s $\rho = -0.74$, $p < 0.001$; Pearson’s $r = -0.68$, $p < 0.001$), computed over individual (language pair × checkpoint) points. We also ensure that this correlation is not merely driven by training dynamics by examining the final checkpoint (step 150,000) in isolation. The relationship remains strong (Spearman’s $\rho = -0.83$, $p < 0.001$; Pearson’s $r = -0.72$, $p < 0.001$), confirming that language pairs which are more separable by form tend to exhibit lower meaning preservation, even in the fully trained model.

\paragraph{Layer-level patterns.} To better understand how these abilities are distributed within the model, Figure~\ref{fig:ABX_across_layers} plots discrimination scores across layers for the final checkpoint. LD is strongest in the lower layers and gradually decreases with depth, reaching a plateau in the upper layers before rising again in the final layer. In contrast, MD starts lower but quickly rises and remains high in the upper layers, but decreases slightly in the last layer. This pattern suggests that earlier layers focus more on identifying the language of the input, while later layers capture its meaning more effectively.

We also find a significant negative correlation between language and meaning discrimination across layers (Spearman’s $\rho = -0.66$, $p < 0.001$; Pearson’s $r = -0.53$, $p < 0.001$). This indicates that, as representations evolve through the network, increases in meaning discrimination are generally accompanied by decreases in language separability. However, the correlation coefficients are weaker than those observed across language pairs, suggesting that this trade-off is not strictly enforced at the layer level (though these values still indicate sizeable effects). Instead, the model exhibits a more flexible allocation of representational capacity across form and meaning over its depth.

\paragraph{Joint checkpoint and layer dynamics.} Figure~\ref{fig:ABX_heatmaps} presents ABX discrimination scores as a function of both checkpoint and layer. LD (left panel) is initially high across most layers but gradually becomes concentrated in the lower layers and the output layer as training advances. In contrast, MD (right panel) improves steadily across all layers, especially in the deeper ones.

\vskip 0.5em

Taken together, these patterns suggest that the model initially relies heavily on language-specific features but gradually shifts toward encoding more abstract, language-invariant semantic structures. Importantly, the two forms of discrimination are not strictly opposed at the layer level. While a trade-off exists, its moderate strength suggests that the model can support both language sensitivity and semantic alignment to some degree simultaneously.

We provide an additional analysis in Appendix~\ref{app:ABX_lang_vs_checkpoint}, showing how both discrimination scores vary across individual languages and training checkpoints.

% This analysis uses the general language and meaning discrimination scores introduced in Section~\ref{sec:abx:data_and_methods}, and reveals meaningful cross-linguistic variation in these dynamics.

\subsection{Discussion}
These findings support the view that pretraining leads to a progressive decoupling of surface form and semantic content. Early in training, language identity is clearly encoded across the model. As training proceeds, this information becomes increasingly concentrated in the lower layers, while deeper layers develop language-invariant semantic representations. This aligns with prior work suggesting that lower layers encode form-related properties, while higher layers abstract away toward more conceptual information \cite{pires2019multilingual, tenney2019bert}. Notably, at convergence, several middle layers appear to support both types of discrimination to a moderate degree, suggesting a partial overlap between structural and semantic signals rather than strict exclusivity.

\section{Correlation of ABX discrimination metrics with linguistic learning}

The ABX results above quantify how form and meaning are structured intrinsically in the embedding space. A natural next question is whether these intrinsic patterns also manifest in extrinsic behavior on standard linguistic tasks. We therefore examine whether variation in ABX discrimination (LD/MD) helps explain differences in probing accuracy and cross-lingual transfer, focusing on monolingual probing and cross-lingual transfer tasks.

We examine whether the model's discrimination patterns relate to linguistic task performance, focusing on monolingual probing and cross-lingual transfer.

\subsection{Experimental Setup}
Following \citet{blevins2022analyzing}, we evaluate both monolingual probing and cross-lingual transfer to test how our ABX discrimination metrics relate to linguistic generalization. We use part-of-speech tagging (POS), named entity recognition (NER), and natural language inference (NLI) as representative tasks. POS and NLI were used in the original analysis; we additionally include NER, which offers a complementary view of lexical-level information and the form–content divide. These tasks span different linguistic levels, from surface form to sentence-level semantics.

All probes are trained independently per language with early stopping on validation loss. We run 6 iterations per setup with different random seeds and report average results. Unless otherwise specified, all experiments use the final XLM-R checkpoint (step 150,000).

\paragraph{Part-of-Speech Tagging (POS)}  

We use Universal Dependencies (UD)~\cite{nivre2020universal}. %, which provides consistent syntactic annotations across languages. 
Monolingual performance is evaluated on all 36 languages from our ABX setup, using standard UD splits. For cross-lingual transfer, we follow \citet{blevins2022analyzing} and use the Parallel UD (PUD) subset at test time, covering 18 languages (Appendix \ref{app:languages}).

\paragraph{Named Entity Recognition (NER)}  

We use WikiAnn~\cite{rahimi2019massively}, providing NER labels in 36 languages. Monolingual evaluation mirrors the POS setup. NER was not included in prior analyses and serves as a new probe of lexical-level representations. For cross-lingual transfer, we use the same 18-language subset used in POS.

\paragraph{Natural Language Inference (NLI)} 
For NLI, we use XNLI~\cite{conneau2018xnli}, a multilingual extension of standard NLI benchmarks. We evaluate both monolingual and cross-lingual performance on the 13 XNLI languages that overlap with our 36-language set.

\subsection{Discrimination Scores and Monolingual Linguistic Probing}\label{sec:disc_and_monolingual_probing}

Following prior work~\cite{blevins2022analyzing}, we find the best checkpoint for probe accuracy varies across tasks and languages (see Appendix~\ref{app:probe_checkpoints})\footnote{We exclude checkpoint 450{,}000 from all analyses due to a training instability, probably due to gradient clipping, that affects both probing and discrimination metrics (see Figure~\ref{fig:ABX_across_checkpoints}).}.

To assess whether language (LD) or meaning discrimination (MD) predict probing performance, we regress POS, NER, and NLI accuracy against each language’s global LD and MD scores, using both the final checkpoint (Last) and the mean across checkpoints (Avg.). For each setting, we fit a multiple linear regression of the form:
\[
\text{Accuracy} \sim \beta_0 + \beta_1 \cdot \text{LD} + \beta_2 \cdot \text{MD} + \epsilon
\]

Table~\ref{tab:regression_summary} summarizes the results\footnote{We also ensure that these effects are not driven by training data size. Language-wise probing accuracy shows no significant correlation with pretraining data quantities (taken from \citet{conneau2020unsupervised}).}. For POS, language discrimination is a robust negative predictor of accuracy, while meaning discrimination shows no significant effect. This suggests that languages which are more easily distinguishable from others (i.e., with higher LD scores) tend to perform worse on syntactic probing tasks, consistent with the idea that strong language-specific encoding may hinder generalization of structural information across languages (see Appendix~\ref{app:ld_vs_probe} for visualization).

\begin{table}[h]
% \centering
\tiny
\begin{tabular}{l l c c l l}

\toprule
\textbf{Setting} & \textbf{Task} & \textbf{Ckpt} & $R^2$ & \textbf{LD Coef (p)} & \textbf{MD Coef (p)} \\
\midrule
Prob. & POS & Avg. & 0.395 & $\mathbf{-2.34}$ \textbf{(}$\mathbf{p < .01}$\textbf{)} & $-0.26$ (n.s.) \\
Prob. & POS & Last & 0.37 & $\mathbf{-1.78}$ \textbf{(}$\mathbf{p < .01}$\textbf{)} & $-0.36$ (n.s.) \\
Prob. & NER & Avg. & 0.085 & $-0.578$ (n.s.)& $-0.074$ (n.s.) \\
Prob. & NER & Last & 0.087 & $-0.584$ (n.s.) & $-0.147$ (n.s.) \\
Prob. & NLI & Avg & 0.275 & $-0.07$ (n.s.) & $-0.12$ (n.s.) \\
Prob. & NLI & Last & 0.224 & $-0.08$ (n.s.) & $-0.14$ (n.s.) 
\\\midrule
CL & POS & Last & 0.324 & $\mathbf{-1.66}$ \textbf{(}$\mathbf{p < .001}$\textbf{)} & $-0.07$ (n.s.) \\
CL & NER & Last & 0.146 & $\mathbf{-0.51}$ \textbf{(}$\mathbf{p < .001}$\textbf{)} & $+0.06$ (n.s.) \\
CL & NLI & Last & 0.3009 & $-0.1$ (n.s.) & $-0.015$ (n.s.) \\
\bottomrule
% \end{tabularx}
\end{tabular}
\caption{
Summary of linear regression results predicting POS and NER accuracy from language discrimination (LD) and meaning discrimination (MD) scores. Each row corresponds to a probing (Prob.) or cross-lingual (CL) evaluation setting.
}
\label{tab:regression_summary}
\end{table}

In contrast, neither LD nor MD significantly predicts performance on NER or XNLI. These tasks may depend less consistently on cross-lingual structural overlap than POS, which could explain the absence of LD as a predictor. While one might expect MD to be predictive, especially for NLI which is explicitly semantic in nature, both tasks may rely on aspects of meaning not well captured by our ABX-based definition of semantic alignment.

% Finally, it is also worth noting that we find no evidence that these effects are driven by training data size. Language-wise probing accuracy shows no significant correlation with pretraining data quantities (taken from \citet{conneau2020unsupervised}).

We also explore whether ABX scores can guide language-specific checkpoint selection, under the hypothesis that lower language discrimination might signal better generalization. We find that LD-based ABX selection improves performance for POS (see Appendix~\ref{appendix:abx-checkpoint-selection} for details).

\subsection{Discrimination Scores and Cross-Lingual Transfer}

We evaluate cross-lingual transfer on POS, NER, and NLI at the final checkpoint.  As originally found by \citet{blevins2022analyzing}, transfer accuracy varies widely across source–target pairs (see Appendix~\ref{appendix:crosslingual_matrix} for detailed heatmaps).
To test whether ABX discrimination explains this variation, we fit linear regression models predicting transfer accuracy from LD and MD scores between language pairs (see Table \ref{tab:regression_summary}). We find that LD is a significant negative predictor for both POS and NER. Neither LD nor MD is predictive of NLI performance. This supports the view that strong language-specific encoding can hinder generalization across languages (see Appendix~\ref{appendix:LD_correl_pos_ner} for visualization).
We also test whether ABX language discrimination can guide source language selection for transfer. While it does not consistently identify the single best source, it often selects competitive candidates and outperforms random baselines (see Appendix~\ref{appendix:abx-source-selection}).

\subsection{Discussion}

A key finding is that language discrimination negatively correlates with POS performance in both monolingual and cross-lingual settings. For NER, LD is predictive only in the cross-lingual setting, suggesting that language-specific encoding affects transfer between languages but has less impact on within-language structure.
Importantly, ABX discrimination scores' interpretation differs slightly between settings. In monolingual probing, LD/MD scores are averaged across all language pairs per language, while cross-lingual transfer uses pair-specific scores for each source–target combination. This finer granularity may help capture transfer-specific effects, explaining why LD predicts cross-lingual NER but not monolingual performance: interference may depend more on the relationship between particular languages than on a language’s overall discriminability. Overall, these results suggest that when a language is highly discriminable from others, its representations may become more isolated, reducing structural sharing and hindering transfer. In the case of POS, this is especially apparent in monolingual probing, where high LD may reflect a failure to encode shared syntactic patterns.

By contrast, MD does not significantly predict downstream accuracy in any task. While one might expect MD to relate to semantically oriented tasks like NLI, success there may depend on higher-level reasoning % or dataset artifacts 
%not well captured 
unaccounted for by our contrastive ABX metric. 
Prior work has also highlighted problematic annotation artifacts, not to mention hypothesis-only biases in the original SNLI dataset from which XNLI was developed, that limit its use for measuring semantic generalization \citep{poliak2018hypothesis, gururangan2018annotation}.

\section{Conclusion and Future Work}

This work introduces ABX-style discrimination metrics for testing how multilingual encoder models discriminate language identity (form) and semantic content (meaning). 
Adapting ABX to text-based multilingual models, we provide a lightweight, interpretable tool for analyzing representational structure.

Applied to XLM-R \cite{conneau2020unsupervised}, our analysis reveals consistent trends across training and depth: language discrimination decreases and concentrates in lower layers, while meaning discrimination increases and stabilizes in deeper ones. This suggests a shift from form-sensitive to meaning-oriented representations during training, without implying a strict trade-off.
We also examine how these metrics relate to downstream performance. Higher language discrimination correlates with lower accuracy on form-sensitive tasks such as POS and NER, while meaning discrimination shows no consistent link, pointing to a possible disconnect between representational alignment and task requirements.
These findings position ABX discrimination as a useful metric for analyzing how multilingual models separate linguistic form from content. They offer a new lens on the evolving structure of multilingual representation spaces and the balance between language-specific and language-invariant information. These metrics could also support practical use cases, such as adaptive checkpoint selection or lightweight diagnostics in multilingual pipelines.

Future work can build on this in several directions. First, discrimination patterns could be linked to typological linguistic features, and it is worth investigating how these typological differences can influence form and content discrimination scores, as previous work has found positive transfer when pairing typologically similar languages \cite{wu2020all}. Second, while we evaluated tasks spanning syntax and semantics (POS, NER, NLI), deeper semantic tasks could better test the role of meaning discrimination. Third, because ABX metrics are architecture-agnostic, they can be applied beyond encoder-only transformers, including decoder-only LLMs and more recent multilingual models, enabling cross-architecture comparisons.
Finally, although ABX does not directly measure representational separation, high discrimination may suggest that a language occupies a distinct subspace. This raises a broader question: how much language sensitivity (i.e., the ability to discriminate languages) can a model have without harming cross-lingual transfer, and how can models balance this trade-off between promoting representational sharing and avoiding interference?

\section*{Limitations}

While our approach provides a detailed analysis of discrimination in multilingual models, it comes with a number of limitations that constrain its generality and suggest directions for future research.

\paragraph{Encoder architecture and single-model scope} Our analysis focuses exclusively on encoder-only architectures, specifically XLM-R. This choice is motivated by the fact that encoder models based on masked language modeling provide stable, structured, layer-wise representations, which are well suited to probing and contrastive analysis. We deliberately restrict our experiments to a single model in order to validate the ABX framework itself, rather than provide a broad multi-model benchmark.  While this makes them a natural starting point for validating our ABX discrimination framework, it remains an open question whether similar dynamics hold for decoder-only or encoder–decoder models, which are trained using autoregressive or sequence-level objectives. Extending our framework to such architectures and to more recent multilingual models is an important direction for future work.

\paragraph{Discrimination vs Separation} Although we distinguish clearly between language and meaning discrimination, we do not explicitly quantify separation in the representation space (e.g., via clustering structure, inter-class variance). Our results suggest that discrimination scores may indirectly reflect separation, but further work is needed to validate this link and to determine whether a model can be discriminative without being structurally partitioned.

\paragraph{Task coverage}
Our evaluation focuses on POS tagging, NER, and NLI, which primarily target syntactic and sentence-level semantic understanding. While these tasks are widely used and informative, they may not capture deeper semantic, pragmatic, or discourse-level capabilities. As a result, the role of meaning discrimination in supporting more abstract or context-sensitive generalization remains an open question.

\paragraph{Cross-linguistic generality vs. language-specific phenomena}
Finally, our analysis examines language and meaning discrimination broadly across multiple languages, but does not investigate the intricacies of specific languages or language families. Languages exhibit unique structural properties, morphological complexity, and semantic nuances that may be represented differently in multilingual models. Future work should explore language-specific discrimination patterns, particularly for typologically diverse languages, to better understand how models encode both universal and language-specific linguistic properties. This would provide insights into representational trade-offs that occur when accommodating multiple languages within a shared parameter space.
% \section*{Acknowledgments}

\bibliography{custom}

\newpage

\input{appendix}

\end{document}

%% file: appendix.tex
\appendix

\section{Illustration of Language and Meaning ABX Setups}\label{appendix:abx-setups}

\begin{figure*}[t]
  \centering
  \begin{subfigure}[t]{0.45\textwidth}
    \includegraphics[trim=5cm 2cm 28cm 1cm,clip,width=\linewidth]{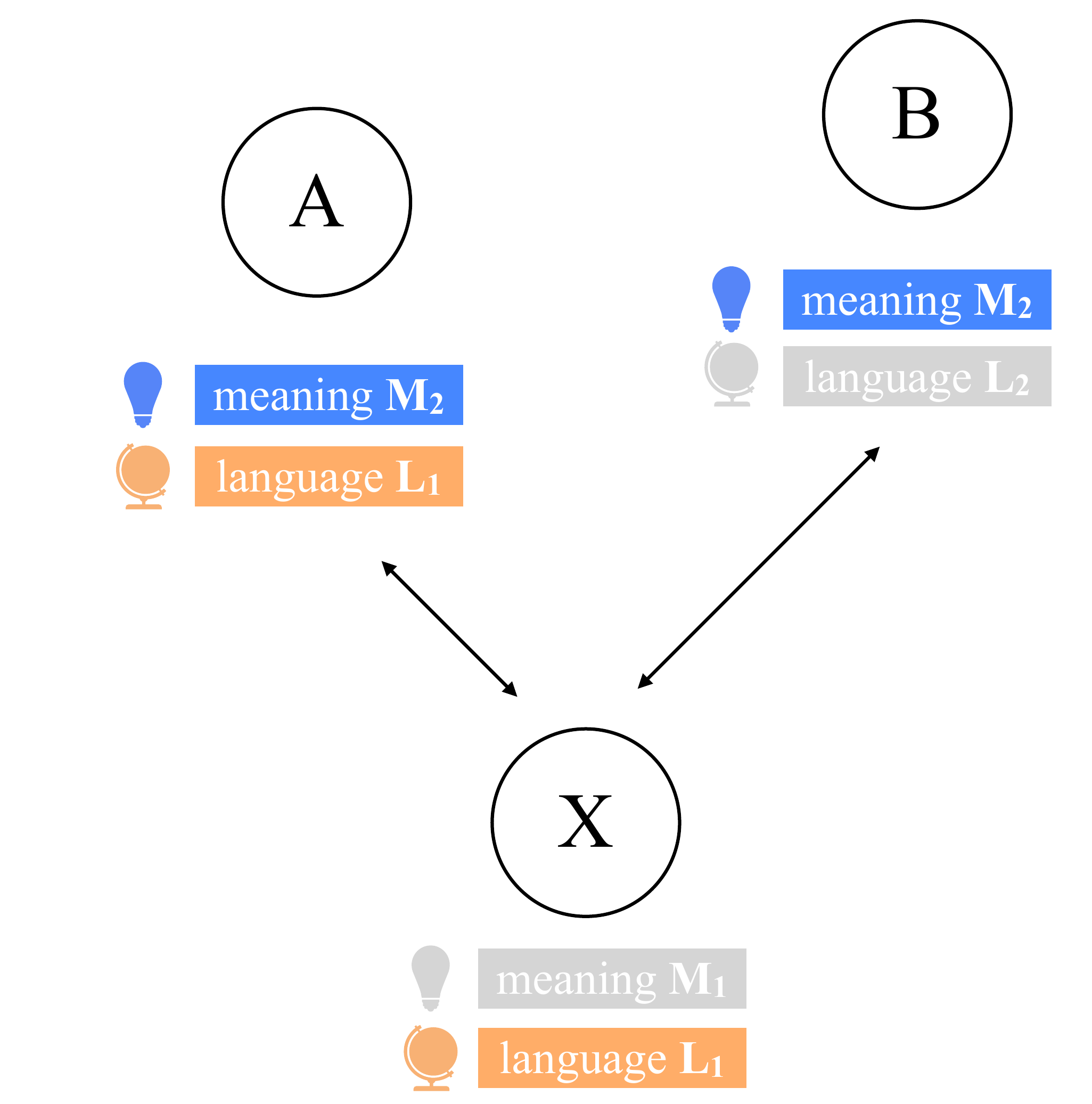}

    \caption{Language Discrimination}
    \label{fig:language_discrimination}
  \end{subfigure}
  \hfill
  \begin{subfigure}[t]{0.45\textwidth}
    \includegraphics[trim=5cm 2cm 28cm 1cm,clip,width=\linewidth]{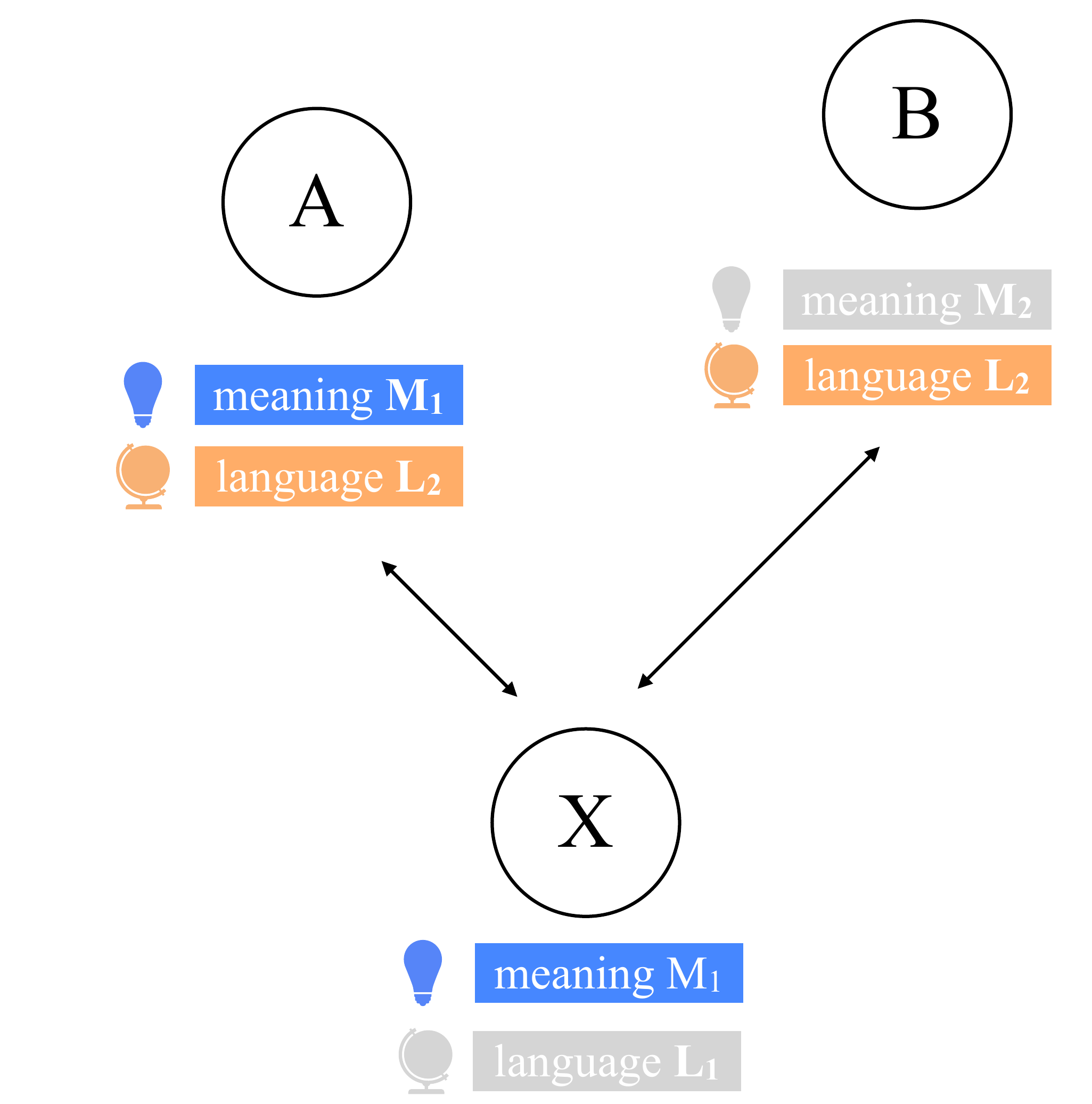}
    \caption{Meaning Discrimination}
    \label{fig:meaning_discrimination}
  \end{subfigure}
  \caption{Illustration of the Language Discrimination (left) and Meaning Discrimination (right) ABX tasks.}
  \label{fig:abx_separate_setups}
\end{figure*}

Figure \ref{fig:abx_separate_setups} illustrates our adaptation of the ABX discrimination paradigm for evaluating multilingual text representations. The figure depicts our two complementary evaluation setups: Language Discrimination (LD) and Meaning Discrimination (MD).

In both setups, we follow a consistent structure where $A$ and $X$ share the variable of interest (the property we want the model to discriminate), while $B$ and $X$ share a control variable (the property we want to control for). Success is measured by whether the model places $X$ closer to $A$ than to $B$ in the embedding space.

For the Language Discrimination task (left panel), the variable of interest is language identity, while meaning serves as the control variable. Specifically, $A$ and $X$ share the same language ($L_1$) but express different meanings, while $A$ and $B$ share the same meaning but are expressed in different languages. When $d(X,A) < d(X,B)$, the model successfully discriminates based on language identity despite semantic differences.

\vspace{0.5 em}
\noindent An example is given below:

\begin{itemize}
    \item \(X\): ``The weather is nice today.'' (English ($L_1$), meaning \(M_1\))
    \item \(A\): ``I need to buy groceries.'' (English ($L_1$), meaning \(M_2\))
    \item \(B\): ``Je dois acheter des provisions.'' (French ($L_2$), meaning \(M_2\): ``I need to buy groceries'')
\end{itemize}

For the Meaning Discrimination task (right panel), the variable of interest is semantic content, while language identity serves as the control variable. Here, $A$ and $X$ share the same meaning ($M_1$) but are expressed in different languages, while $A$ and $B$ share the same language ($L_2$) but express different meanings. When $d(X,A) < d(X,B)$, the model successfully discriminates based on semantic similarity across languages despite surface form differences.

\vspace{0.5 em}
\noindent Here is an example for the MD task:

\begin{itemize}

    \item \(X\): ``The weather is nice today.'' (English ($L_1$), meaning \(M_1\))
    \item \(A\): ``La météo est bonne aujourd'hui'' (French ($L_2$), meaning \(M_1\): ``The weather is nice today'')
    \item \(B\): ``Je dois acheter des provisions.'' (French ($L_2$), meaning \(M_2\): ``I need to buy groceries'')
\end{itemize}

This systematic approach allows us to isolate specific properties in multilingual representations by controlling for potential confounding factors. The ABX score for each task reflects the proportion of triplets where the model correctly places items sharing the variable of interest closer together than those sharing only the control variable, providing a direct measure of how the model structures linguistic information along these dimensions.

\section{Pseudocode for ABX Evaluation}
\label{appendix:pseudocode}

We provide pseudocode for the ABX evaluation pipeline. Triplet sampling specifies how $(A,B,X)$ items are constructed for the two tasks: Language Discrimination (LD) and Meaning Discrimination (MD). 
Scoring then determines whether $X$ is closer to $A$ or $B$  in representation space, using cosine distance at a fixed layer $l$ of the model. The final ABX score is obtained by averaging over  many sampled triplets.

\paragraph{Triplet sampling.}
Algorithm~\ref{alg:ld} describes how triplets are drawn for  Language Discrimination, and Algorithm~\ref{alg:md} for Meaning Discrimination. The corresponding setups are 
illustrated in Figure~\ref{fig:abx_separate_setups} 
(Appendix~\ref{appendix:abx-setups}), which provides 
examples of how $A$, $B$, and $X$ are selected in practice.\\

\vspace{0.2em}

\noindent\textit{Remark.} In practice, ABX scores are estimated from a large random sample of triplets rather than all possible  combinations. The number of triplets controls stability: higher  sample sizes give lower variance at the cost of computation. 
In our experiments, we fix the number of triplets per condition  ($\approx$\,100k) to ensure reproducibility and comparability across checkpoints. 
In a robustness check, we recomputed ABX scores over five different random subsamples (final checkpoint). The average standard deviation was below 0.01, confirming that our results are highly stable with 
respect to triplet sampling.

\begin{algorithm}[H]
\caption{Sampling ABX triplets for Language Discrimination (LD)}
\label{alg:ld}
\begin{algorithmic}[1]
\Require Parallel corpus with aligned positions $p$ for languages $\ell_1,\ell_2$
\For{each language pair $(\ell_1,\ell_2)$}
  \State Pick $A$ in $\ell_1$ at position $p_A$
  \State Pick $B$ in $\ell_2$ at the same position $p_A$ \Comment{translation of $A$; same meaning}
  \State Pick $X$ in $\ell_1$ at some $p_X \neq p_A$ \Comment{same language as $A$; different meaning}
  \State Emit $(A,B,X)$
  \State Repeat with $(\ell_1,\ell_2)$ swapped
  \State Repeat until desired sample size
\EndFor
\end{algorithmic}
\end{algorithm}

\begin{algorithm}[H]
\caption{Sampling ABX triplets for Meaning Discrimination (MD)}
\label{alg:md}
\begin{algorithmic}[1]
\Require Parallel corpus with aligned positions $p$ for languages $\ell_1,\ell_2$
\For{each language pair $(\ell_1,\ell_2)$}
  \State Pick $X$ in $\ell_1$ at position $p_X$
  \State Pick $A$ in $\ell_2$ at the same position $p_X$ \Comment{translation of $X$; same meaning}
  \State Pick $B$ in $\ell_2$ at some $p_B \neq p_X$ \Comment{same language as $A$; different meaning}
  \State Emit $(A,B,X)$
  \State Repeat with $(\ell_1,\ell_2)$ swapped
  \State Repeat until desired sample size
\EndFor
\end{algorithmic}
\end{algorithm}

\paragraph{ABX discrimination scoring.}
Algorithm~\ref{alg:scoring} shows how a score is computed 
from each triplet. A score of 1 is assigned if $X$ is closer to $A$ 
than to $B$, 0 if the opposite, and 0.5 in case of a tie. 
Averaging across all triplets gives the ABX score for the condition.

\begin{algorithm}[H]
\caption{ABX discrimination scoring}
\label{alg:scoring}
\begin{algorithmic}[1]
\Require Triplets $(A,B,X)$, pretrained model $M$ at checkpoint $c$, tokenizer $T$, layer $l$
\State Define $d(u,v) = 1 - \frac{u \cdot v}{\lVert u\rVert \lVert v\rVert}$ \Comment{cosine distance}
\For{each triplet $(A,B,X)$}
  \State Encode $A,B,X$ to embeddings $e_A,e_B,e_X$ (mean over subword states at layer $l$)
  \State Compute $d_{AX} \gets d(e_A,e_X)$, \quad $d_{BX} \gets d(e_B,e_X)$
  \State $s \gets \begin{cases}
    1 & \text{if } d_{AX} < d_{BX} \\
    0 & \text{if } d_{AX} > d_{BX} \\
    0.5 & \text{otherwise}
  \end{cases}$
\EndFor
\State \Return mean ABX score across all triplets
\end{algorithmic}
\end{algorithm}

\section{Languages Used in Evaluations}
\label{app:languages}

Table~\ref{tab:languages} lists all languages selected for our different evaluations, including ABX discrimination tasks and probing tasks. The selection covers a wide range of language families, scripts, and typological characteristics.

\begin{table}[h]
\centering
\resizebox{\columnwidth}{!}{% Scale the table to fit within the text width
\begin{tabular}{llccccccccc}
\toprule
\textbf{Code} & \textbf{Language} &  \textbf{ABX} & \multicolumn{2}{c}{\textbf{POS}} & \multicolumn{2}{c}{\textbf{NER}} & \multicolumn{2}{c}{\textbf{NLI}} \\
 \cmidrule(lr){4-5} \cmidrule(lr){6-7} \cmidrule(lr){8-9}
& &  & mono. & CL & mono. & CL & mono. & CL \\
\midrule
ar & Arabic & \checkmark & \checkmark & \checkmark  &\checkmark  &\checkmark  &\checkmark  &\checkmark  \\
bg & Bulgarian & \checkmark & \checkmark & &\checkmark  &\checkmark  & \checkmark & \checkmark \\
ca & Catalan &\checkmark  &\checkmark  & &\checkmark  & & & \\
cs & Czech &\checkmark  &\checkmark  &\checkmark  &\checkmark  &\checkmark  & & \\
da & Danish & \checkmark &\checkmark  & &\checkmark  & & & \\
de & German & \checkmark & \checkmark & \checkmark & \checkmark & \checkmark & \checkmark & \checkmark \\
el & Greek & \checkmark &\checkmark  & & \checkmark &\checkmark  & \checkmark &\checkmark  \\
en & English & \checkmark & \checkmark & \checkmark & \checkmark & \checkmark & \checkmark & \checkmark \\
es & Spanish & \checkmark & \checkmark & \checkmark & \checkmark & \checkmark & \checkmark & \checkmark \\
et & Estonian & \checkmark &\checkmark  & &\checkmark  & & & \\
fa & Persian &\checkmark  &\checkmark  & & \checkmark & & & \\
fi & Finnish &\checkmark  &\checkmark  & \checkmark & \checkmark &\checkmark  & & \\
fr & French & \checkmark & \checkmark & \checkmark & \checkmark & \checkmark & \checkmark & \checkmark \\
he & Hebrew &\checkmark  &\checkmark  & &\checkmark  & & & \\
hi & Hindi &\checkmark  &\checkmark  &\checkmark  &\checkmark  &\checkmark  & \checkmark &\checkmark  \\
hr & Croatian & \checkmark &\checkmark  & &\checkmark  & & & \\
hu & Hungarian &\checkmark  &\checkmark  & &\checkmark  & & & \\
is & Icelandic &\checkmark  &\checkmark  & \checkmark &\checkmark  &\checkmark  & & \\
it & Italian & \checkmark & \checkmark & \checkmark & \checkmark & \checkmark &  &  \\
ja & Japanese &\checkmark  &\checkmark  &\checkmark  &\checkmark  &\checkmark  & & \\
ko & Korean &\checkmark  &\checkmark  &\checkmark  &\checkmark  &\checkmark  & & \\
lv & Latvian &\checkmark  &\checkmark  & &\checkmark  & & & \\
nl & Dutch &\checkmark  &\checkmark  & &\checkmark  & & & \\
pl & Polish &\checkmark  &\checkmark  &\checkmark  &\checkmark  &\checkmark  & & \\
pt & Portuguese &\checkmark  &\checkmark  &\checkmark  &\checkmark  &\checkmark  & & \\
ro & Romanian &\checkmark  &\checkmark  & &\checkmark  & & & \\
ru & Russian &\checkmark  &\checkmark  & \checkmark &\checkmark  &\checkmark  &\checkmark  & \checkmark \\
sk & Slovak &\checkmark  &\checkmark  & &\checkmark  & & & \\
sl & Slovenian &\checkmark  &\checkmark  & &\checkmark  & & & \\
sr & Serbian &\checkmark  &\checkmark  & &\checkmark  & & & \\
sv & Swedish &\checkmark  &\checkmark  &\checkmark  &\checkmark  & \checkmark &  & \\
tr & Turkish &\checkmark  &\checkmark  &\checkmark  & \checkmark & \checkmark &\checkmark  & \checkmark \\
uk & Ukrainian & \checkmark &\checkmark  & & \checkmark & & & \\
ur & Urdu & \checkmark &\checkmark  & & \checkmark &\checkmark  &\checkmark  & \checkmark \\
vi & Vietnamese &\checkmark  &\checkmark  & &\checkmark  & &\checkmark  &\checkmark  \\
zh & Chinese &\checkmark  &\checkmark  &\checkmark  &\checkmark  &\checkmark  &\checkmark  &\checkmark  \\
\bottomrule
\end{tabular}
}
\caption{Languages and related ISO codes used in discrimination evaluations (ABX), and probing tasks (mono for monolingual probing and CL for cross-lingual).  A checkmark indicates the language is used in that task/subset.}
\label{tab:languages}
\end{table}

\section{Correlation Analysis Between Language and Meaning Discrimination in XLM-R}\label{appendix:correl_md_ld}

This appendix provides additional analysis on the relationship between language discrimination (LD) and meaning discrimination (MD) in our model.

We observe a strong overall negative correlation between LD and MD across all language pairs and checkpoints (Spearman’s $\rho = -0.74$, $p < 0.001$; Pearson’s $r = -0.68$, $p < 0.001$), computed at the (language pair × checkpoint) level. This suggests that, throughout training, language pairs that are more separable in form tend to be less effective in preserving semantic structure.

\begin{figure}[h]
    \centering
    \includegraphics[width=1\linewidth]{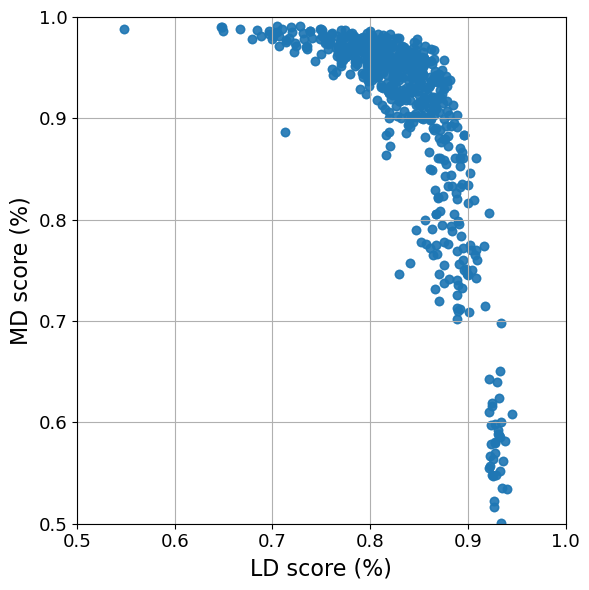}
    \caption{Scatterplot showing the relationship between language discrimination (x-axis) and meaning discrimination (y-axis) scores for all language pairs at checkpoint 150,000 (last). Each point represents a language pair.}
    \label{fig:scatt-corr-ld-md}
\end{figure}

To verify that this effect is not simply an artifact of training progression, we examine the relationship at the final checkpoint (step 150,000) alone. The inverse correlation persists with even greater magnitude (Spearman’s $\rho = -0.83$, $p <0.001$; Pearson’s $r = -0.72$, $p < 0.001$), confirming that the tradeoff between language and meaning discrimination remains pronounced even in the fully trained model. Figure~\ref{fig:scatt-corr-ld-md} visualizes this relationship: the scatterplot reveals a clear monotonic trend, with almost no high–high co-occurrence (i.e., no language pairs simultaneously scoring high on both MD and LD), which supports the interpretation of a representational tradeoff.

% We further analyze the dynamics of this relationship across training by computing correlation coefficients at each checkpoint (Figure~\ref{fig:spearman-pearson-md-ld-corr}). Spearman’s correlation remains consistently strong and statistically significant across all training stages, suggesting a stable monotonic inverse relationship. Pearson’s correlation, while also consistently negative, varies in magnitude but remains significant as training progresses, indicating that the relationship is not only ordinal but approximately linear in later stages. 

% \begin{figure}[h]
%   \centering
%   \includegraphics[width=\linewidth]{images/evolution_spearman_ld_md.png}
%   \vspace{0.5em}
%   \includegraphics[width=\linewidth]{images/evolution_pearson_ld_md.png}
%   \caption{Evolution of Spearman (top) and Pearson (bottom) correlation coefficients between language and meaning discrimination scores across training checkpoints. Statistically significant correlations (p < 0.05) are highlighted.}
%   \label{fig:spearman-pearson-md-ld-corr}
% \end{figure}

\begin{figure}[h]
  \centering
  \includegraphics[width=\linewidth]{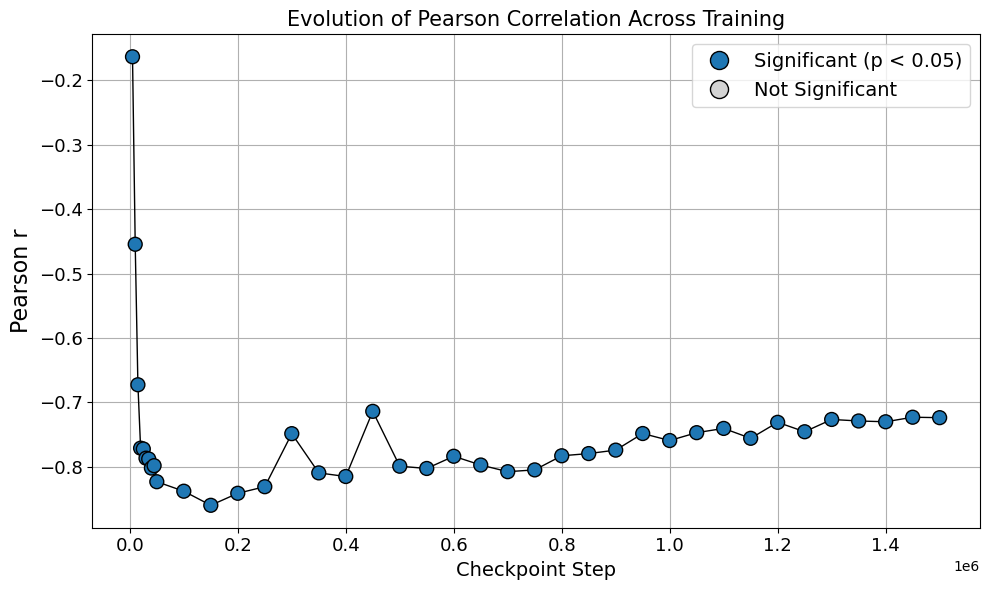}
  \caption{Evolution of Pearson correlation coefficients between language and meaning discrimination scores across training checkpoints. Statistically significant correlations ($p < 0.05$) are highlighted.}
  \label{fig:pearson-md-ld-corr}
\end{figure}

We further analyze the dynamics of this relationship across training by computing correlation coefficients at each checkpoint (Figure~\ref{fig:pearson-md-ld-corr}). Pearson’s correlation remains consistently negative and statistically significant across all training stages, indicating that the relationship between LD and MD is approximately linear throughout training.

\section{Discrimination scores across languages and checkpoints}\label{app:ABX_lang_vs_checkpoint}

\begin{figure*}[t]
  \includegraphics[width=0.48\linewidth]{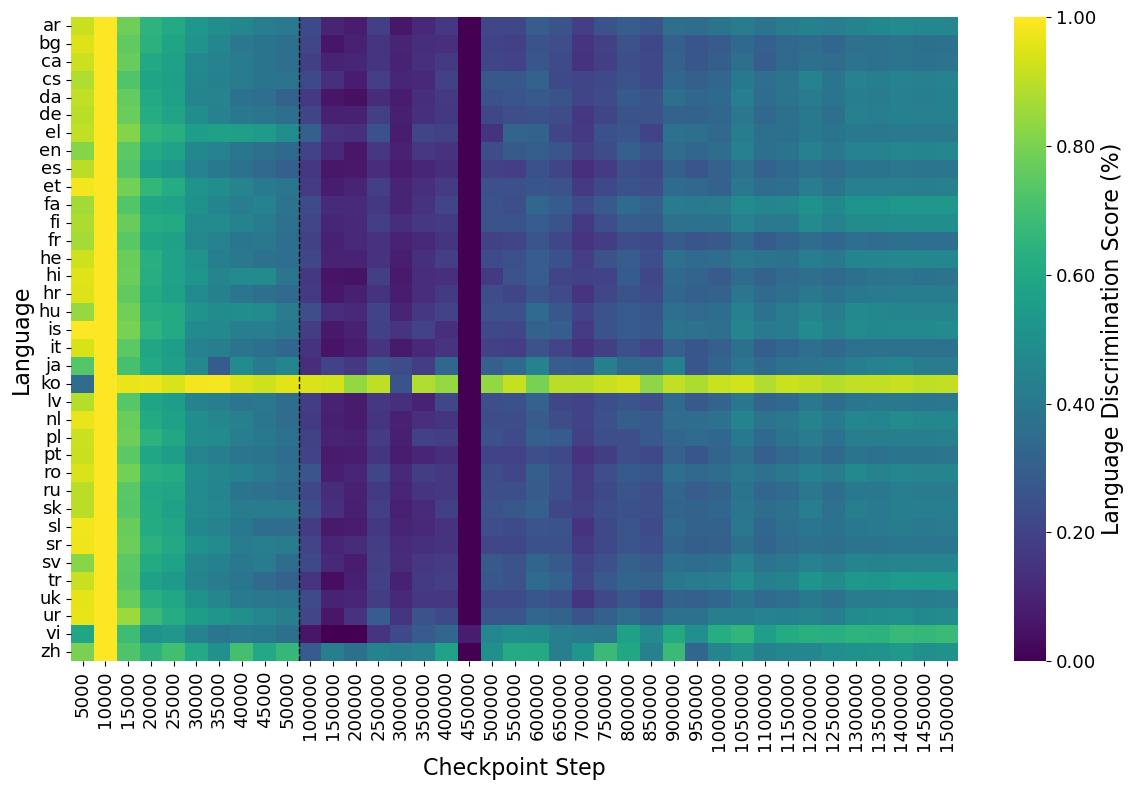} \hfill
  \includegraphics[width=0.48\linewidth]{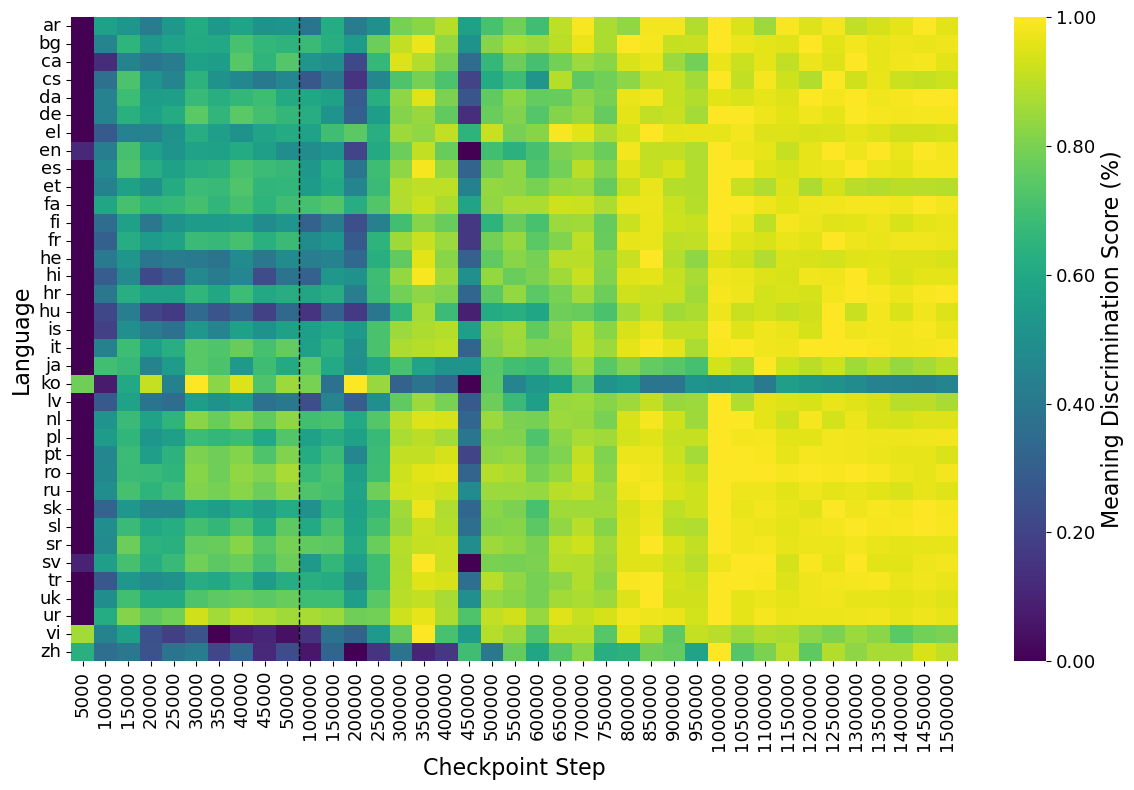}
  \includegraphics[width=0.48\linewidth]{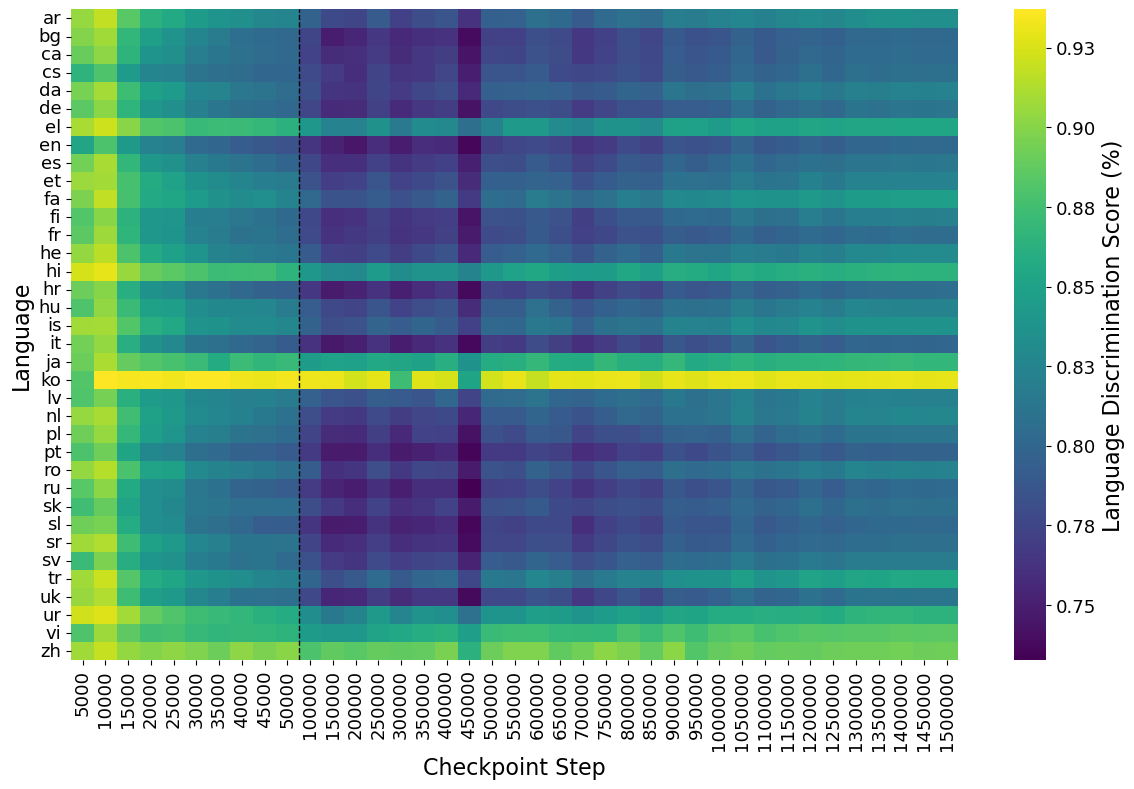} \hfill
  \includegraphics[width=0.48\linewidth]{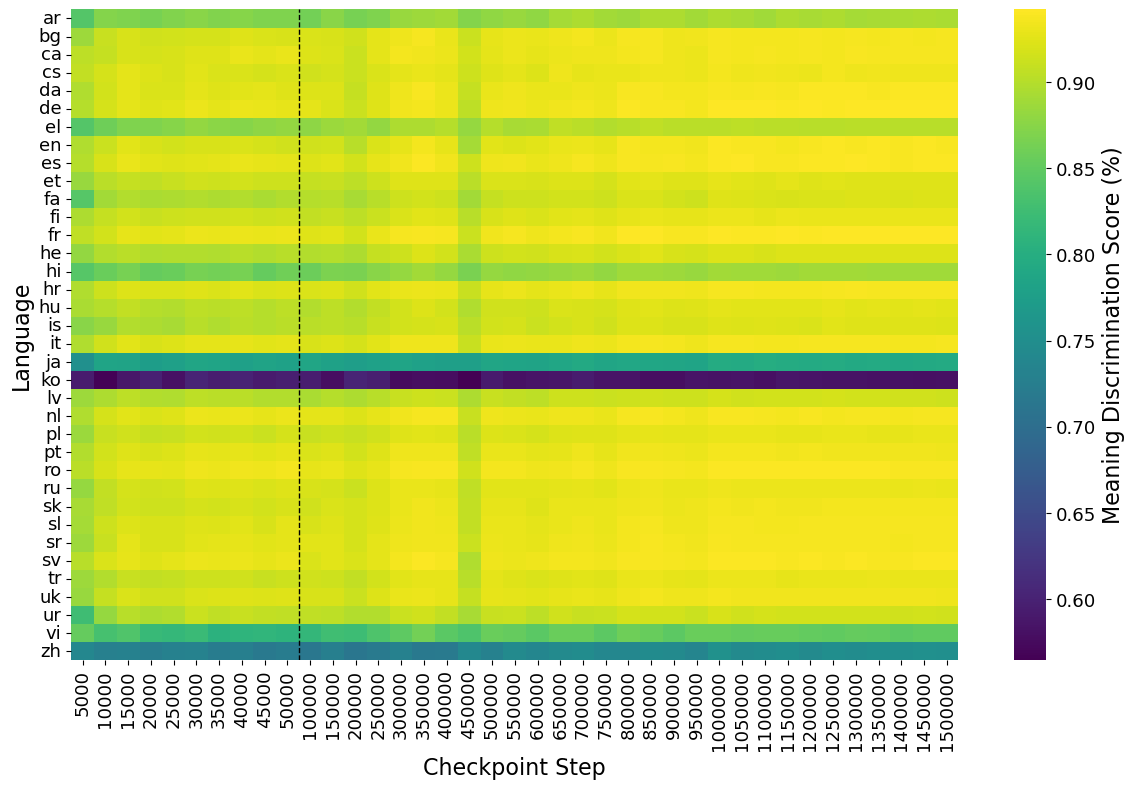}
  \caption{Heatmaps showing the evolution of language discrimination (left) and meaning discrimination (right) \emph{(averaged across layers)} across checkpoints. Scores are normalized per language to highlight the differences across checkpoints (top row). We also provide the non-normalized scores (bottom row). Each heatmap's row represents a language, and each column a checkpoint step. Bright regions indicate high relative discrimination at that training stage for the given language.}
  \label{fig:ABX_lang_vs_checkpoint}
\end{figure*}

\begin{figure*}[t]
  \includegraphics[width=0.48\linewidth]{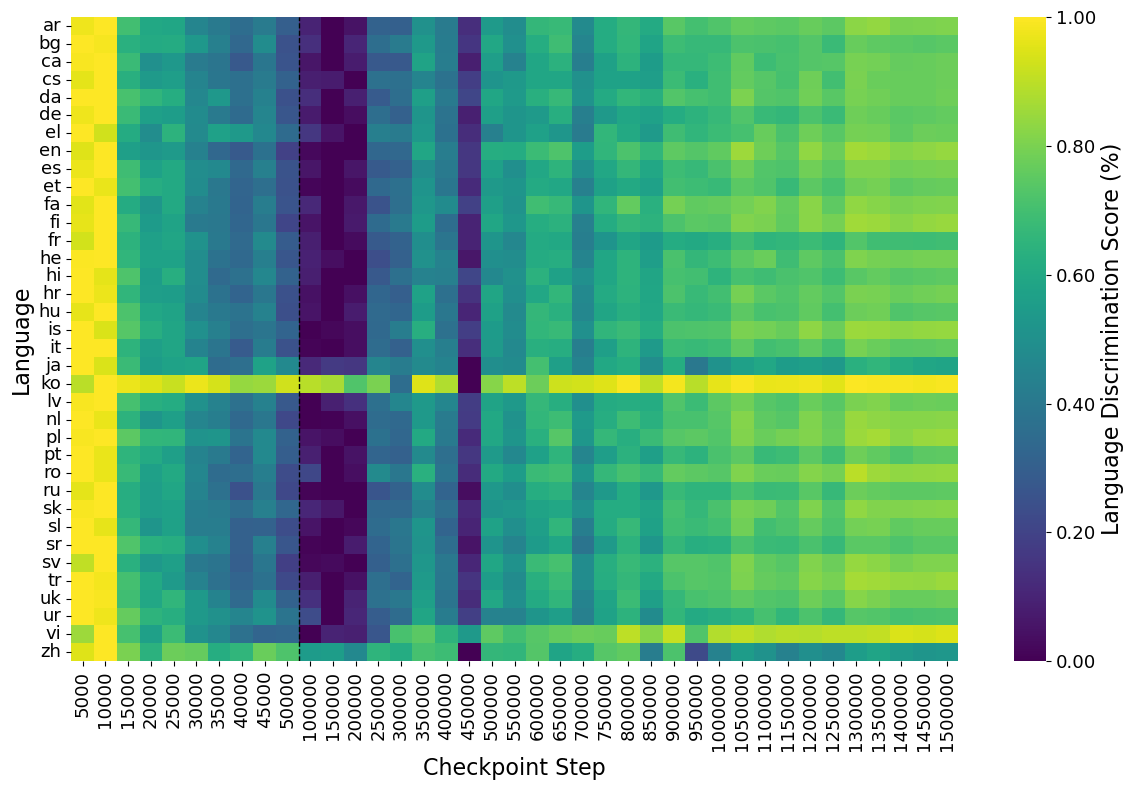} \hfill
  \includegraphics[width=0.48\linewidth]{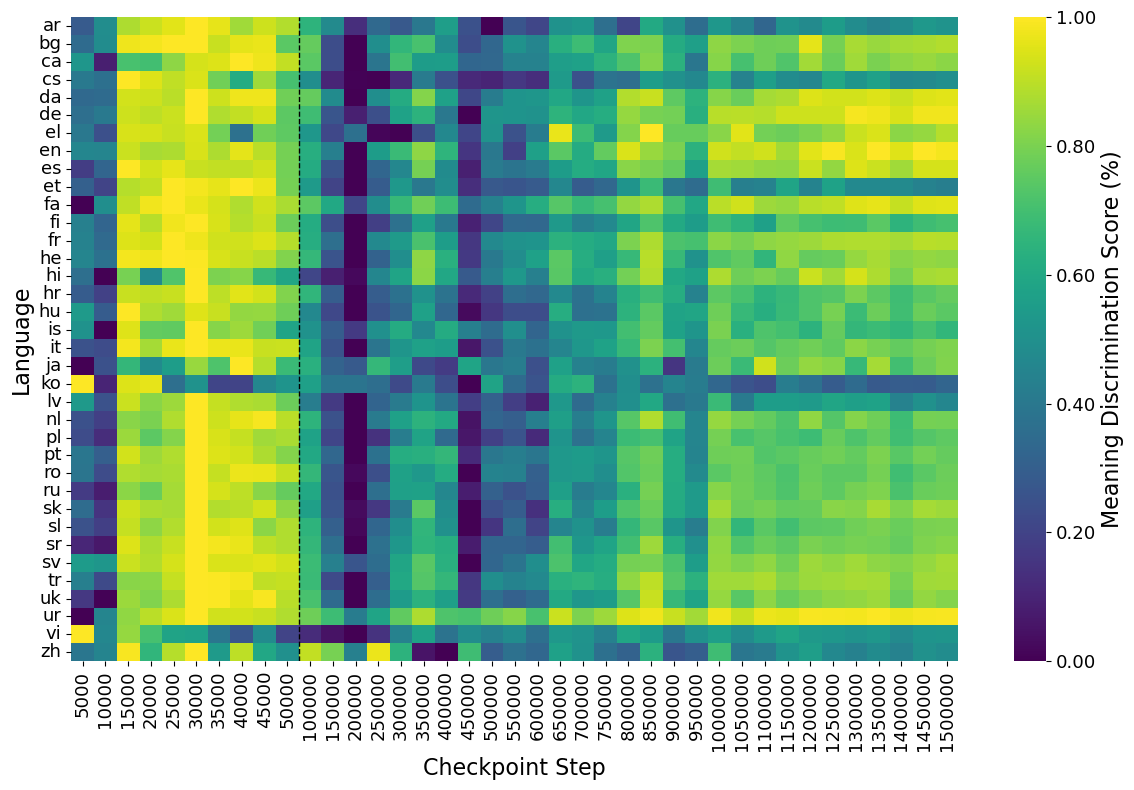}
    \includegraphics[width=0.48\linewidth]{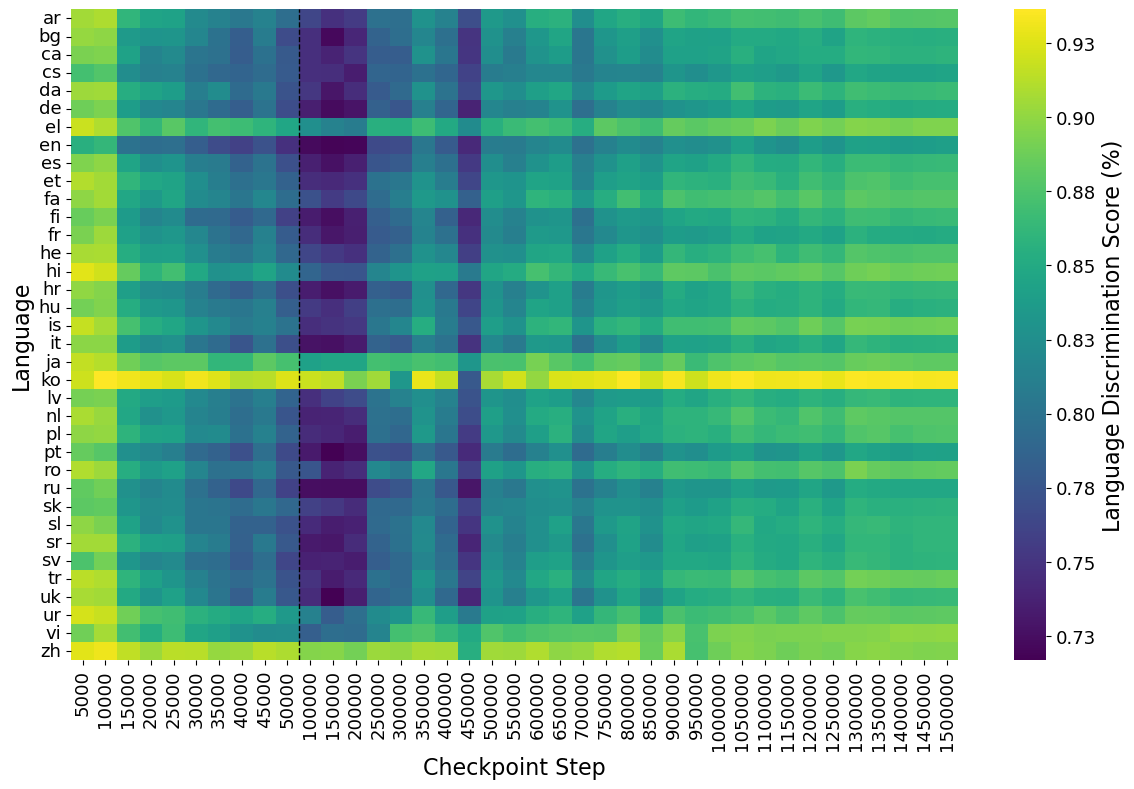} \hfill
  \includegraphics[width=0.48\linewidth]{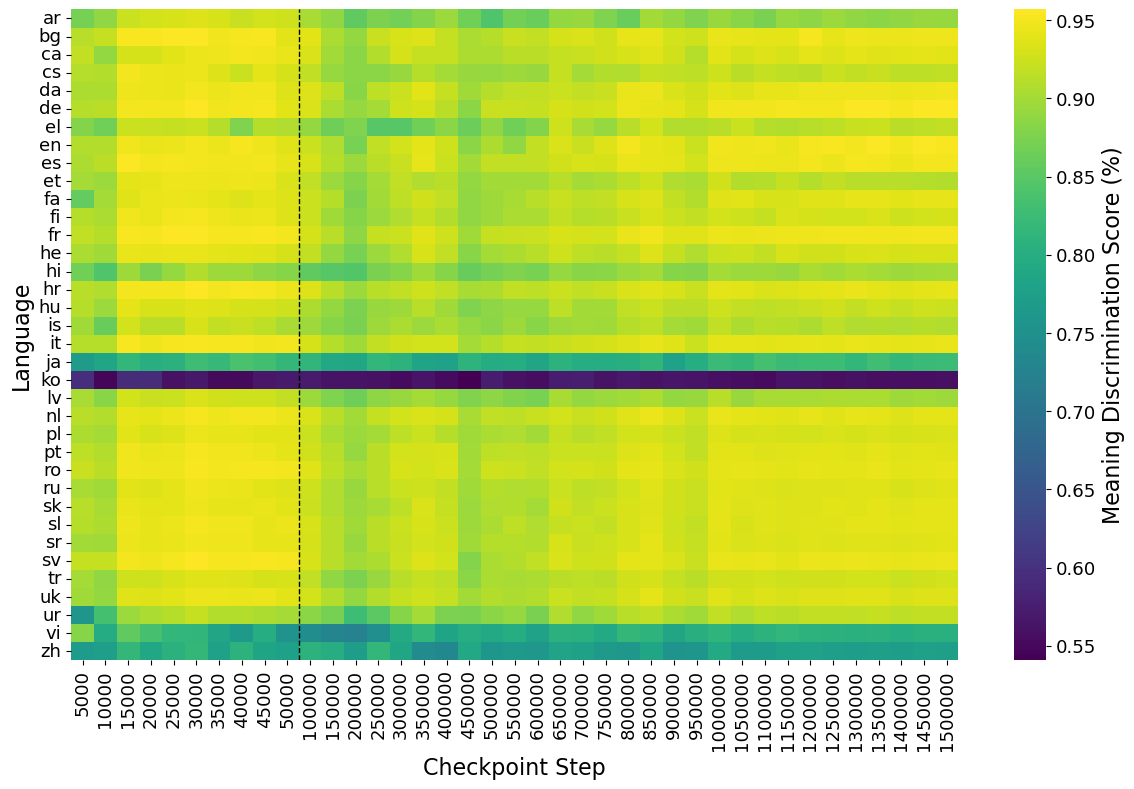}
  \caption{Heatmaps showing the evolution of language discrimination (left) and meaning discrimination (right) on \emph{the last layer}, across checkpoints. Scores are normalized per language to highlight the differences across checkpoints (top row). We also provide the non-normalized scores (bottom row). Each heatmap's row represents a language, and each column a checkpoint step. Bright regions indicate high relative discrimination at that training stage for the given language. }
  \label{fig:ABX_lang_vs_checkpoint_lastlayer}
\end{figure*}

To examine how discrimination evolves at the individual language level, we present heatmaps of language and meaning discrimination scores across checkpoints (Figure~\ref{fig:ABX_lang_vs_checkpoint}). Scores are normalised per language to highlight relative changes over time. For language discrimination (left), we observe a sharp decline during early training steps for most languages, followed by a partial recovery. However, the timing and extent of this rebound varies across languages, suggesting that some retain language-specific features more robustly. In contrast, meaning discrimination (right) increases steadily for all languages, but again at different rates, with certain languages benefiting earlier from semantically structured representations. These differences may reflect both linguistic factors and data resource disparities. Additional views of final-layer behavior are included in Figure \ref{fig:ABX_lang_vs_checkpoint_lastlayer}.

\section{Checkpoint-wise Probe Accuracy}
\label{app:probe_checkpoints}

\begin{figure*}[h]
  \includegraphics[width=0.48\linewidth]{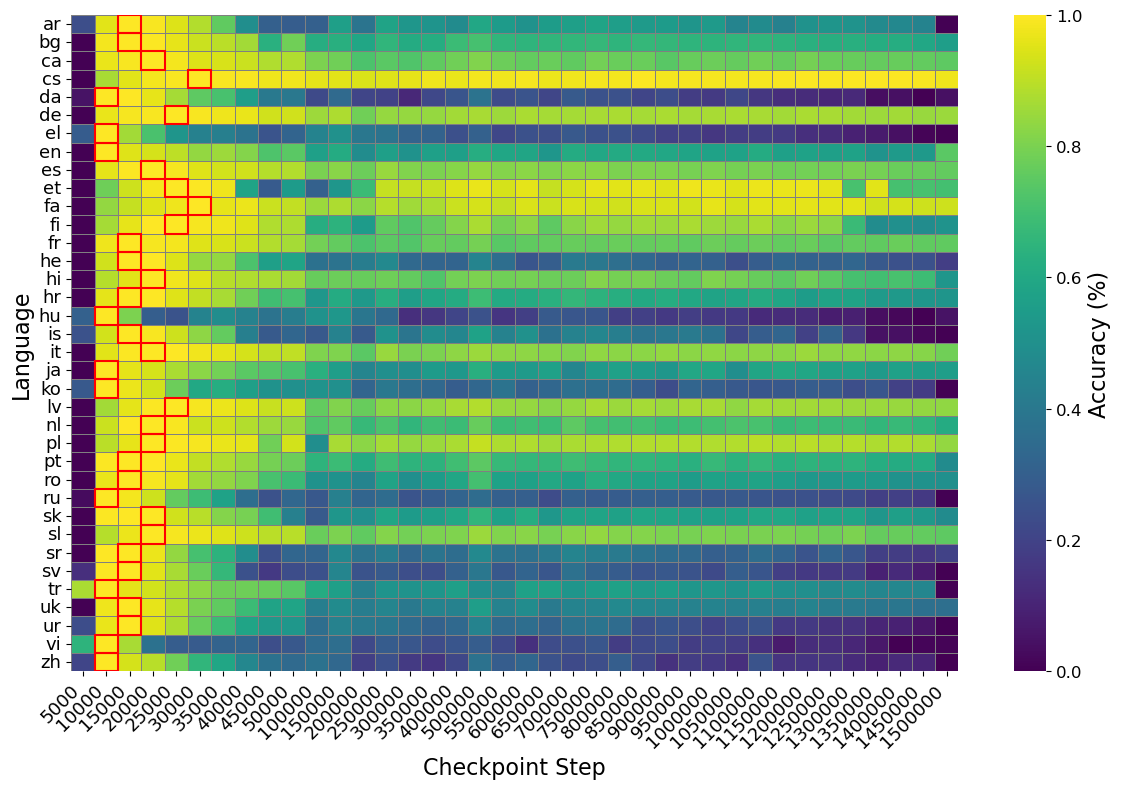} \hfill
  \includegraphics[width=0.48\linewidth]{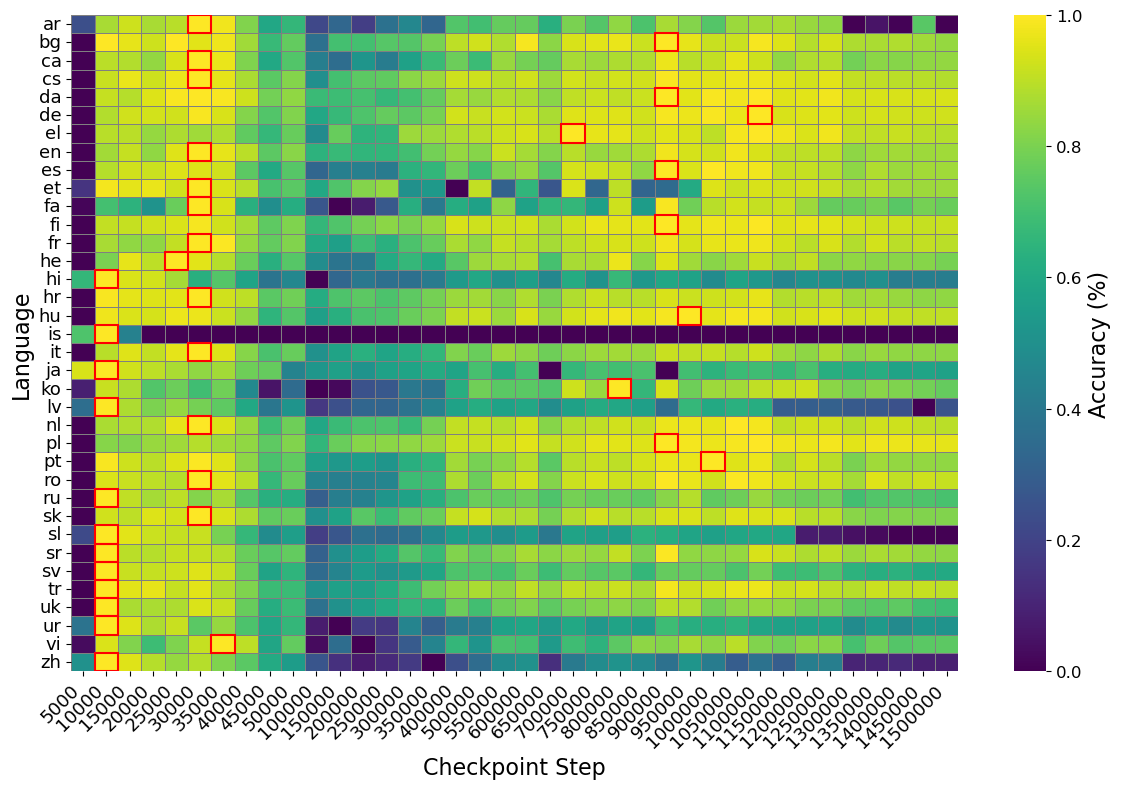}

  \includegraphics[width=0.48\linewidth]{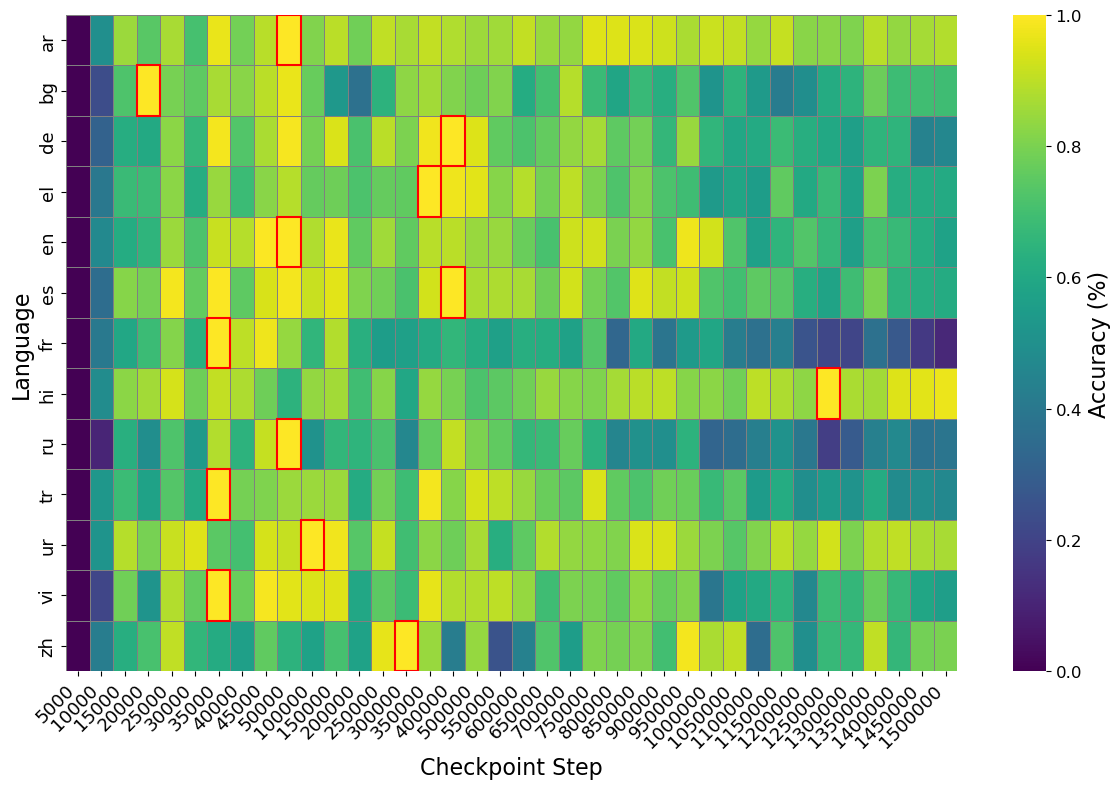}

  \caption{Checkpoint-wise probe accuracy across languages for POS (top left), NER (right), and NLI (bottom left) normalized per language. Each row corresponds to a language, and red boxes mark the checkpoint at which that language reaches peak accuracy for the probing task. Lighter regions mean higher accuracy scores.}
  \label{fig:probe_heatmaps_appendix}
\end{figure*}

Figure~\ref{fig:probe_heatmaps_appendix} shows per-language probe accuracy across checkpoints for POS, NER and NLI highlighting the variability in when each language reaches its peak performance.

\section{Additional Probing Analyses}\label{app:ld_vs_probe}

% Figure~\ref{fig:probe_vs_abx} shows the negative relationship between ABX language discrimination and POS/NER accuracy across languages. Higher language discrimination scores are associated with lower probing performance, consistent with the idea that strong language-specific encoding may limit generalization.

\begin{figure}[h]
  \centering

    \includegraphics[width=0.8\linewidth]{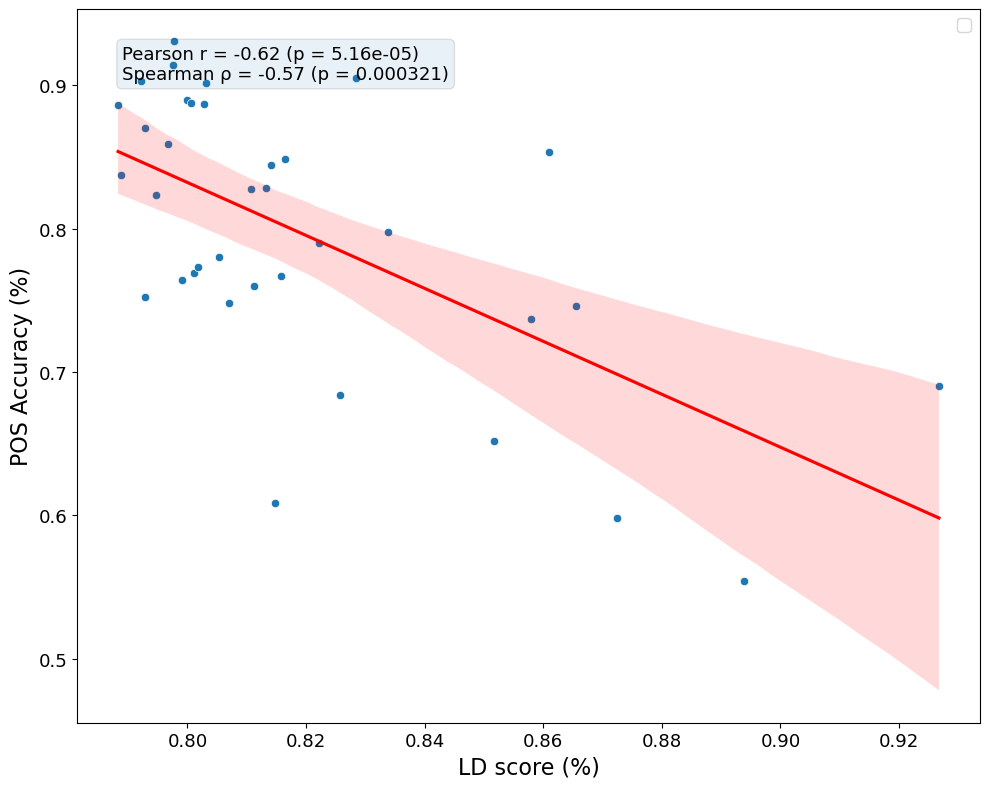}

  \caption{Relationship between ABX-based Language Discrimination scores and downstream probing POS accuracy, averaged across checkpoints.
Each point represents a single evaluation language. The x-axis shows how well the model distinguishes that language from others (higher = more discriminable), while the y-axis shows its average performance on the downstream task. Red lines indicate linear regression fits with shaded 95\% confidence intervals.}
  \label{fig:probe_vs_abx}
\end{figure}

Figure~\ref{fig:probe_vs_abx} shows the negative relationship between ABX language discrimination and POS accuracy across languages. Higher language discrimination scores are associated with lower probing performance, consistent with the idea that strong language-specific encoding may limit generalization.

\section{ABX-Guided Checkpoint Selection}
\label{appendix:abx-checkpoint-selection}

Given that language discrimination is a strong global predictor of probing accuracy for POS, we ask whether ABX scores can serve as lightweight, unsupervised heuristics for language-specific checkpoint selection. Specifically, we evaluate whether selecting, for each language, the checkpoint with minimal LD brings the model closer to its optimal performance, compared to using the final checkpoint uniformly.

We compare probing accuracy at the ABX-selected checkpoint to that at the final training step, measuring their respective distances from each language’s best-performing checkpoint.

% Results are mixed. For POS,

ABX-guided selection yields a closer match to the best checkpoint in 29 out of 36 languages, with a mean improvement of 0.034 ± 0.048, and a Wilcoxon signed-rank test confirming significance over choosing the final checkpoint ($p < 0.001$). This suggests that LD dynamics during training can inform language-specific model selection, particularly when the final checkpoint is suboptimal. 

\begin{figure*}[h]
    \centering

        \includegraphics[width=\textwidth]{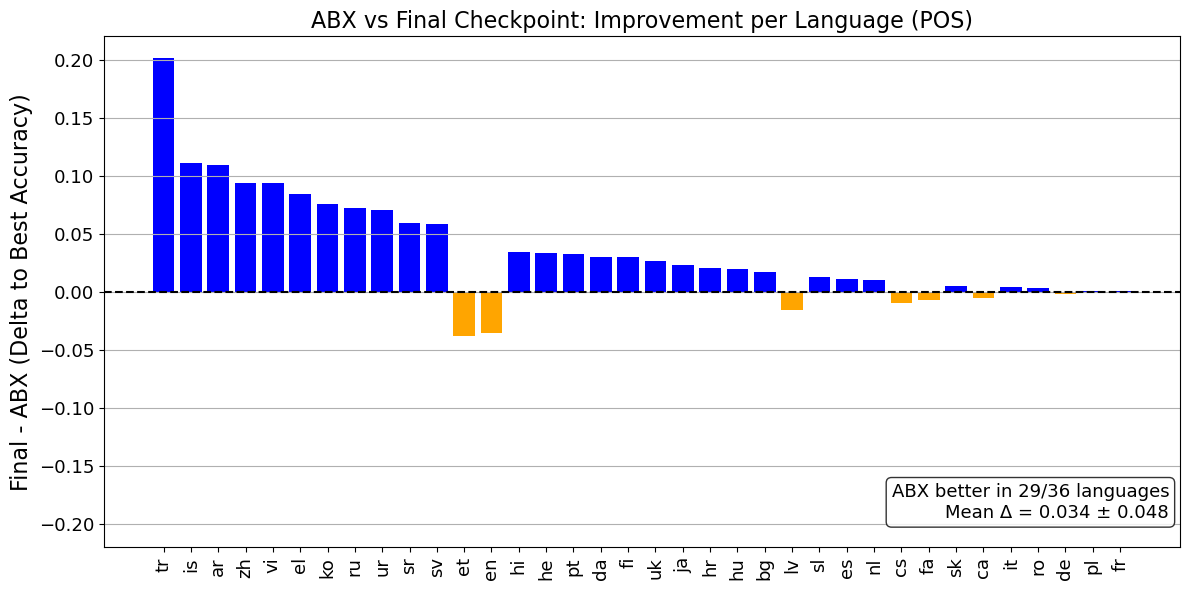}
        \caption{POS}
        \label{fig:abx_vs_final_pos}

    \caption{Difference in performance gap to the best checkpoint for each language, comparing ABX-selected (lowest LD ABX) vs. final checkpoints for the POS task. Bars show the difference in delta (Final - ABX); positive values indicate that the ABX-selected checkpoint is closer to the best-performing one (i.e., smaller gap to optimal accuracy).}
    \label{fig:abx_vs_final_deltas}
\end{figure*}

These patterns are visualised in Figure~\ref{fig:abx_vs_final_deltas}, which presents per-language deltas. For each language, we compute:
\[
\Delta = \text{Final} - \text{ABX}
\]
where positive values indicate that ABX selection yields a checkpoint closer to the best-performing one. Bars are sorted by the absolute delta, highlighting languages with the largest impact.

\section{Cross-Lingual Transfer Accuracy Matrices}
\label{appendix:crosslingual_matrix}

\begin{figure*}[h]
  \includegraphics[width=0.48\linewidth]{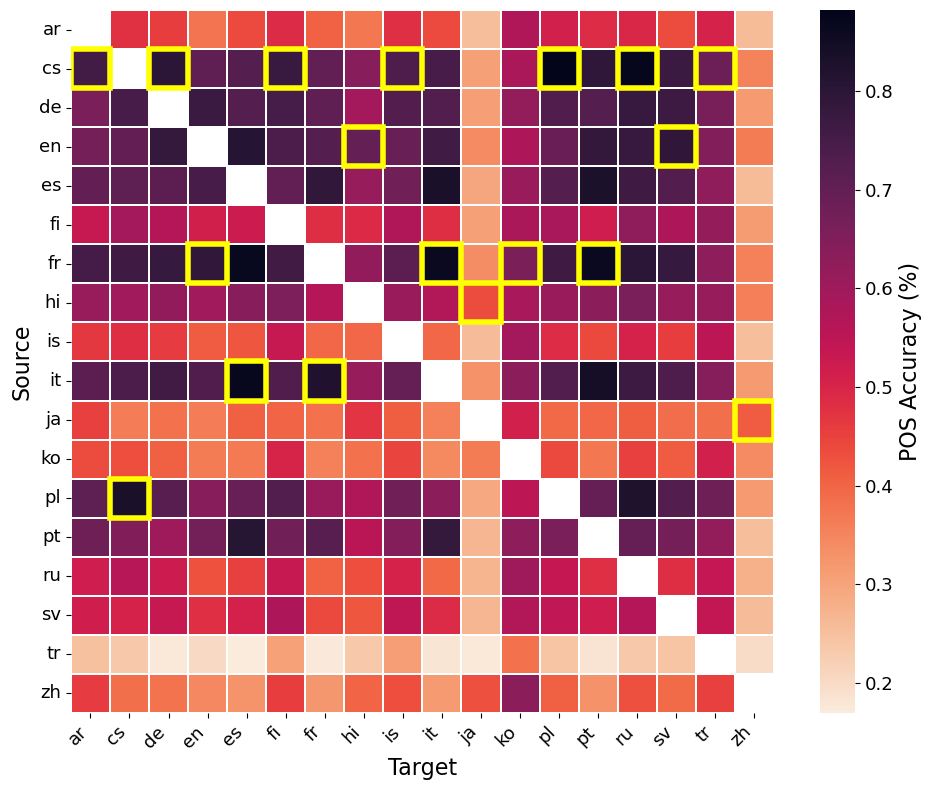} \hfill
  \includegraphics[width=0.48\linewidth]{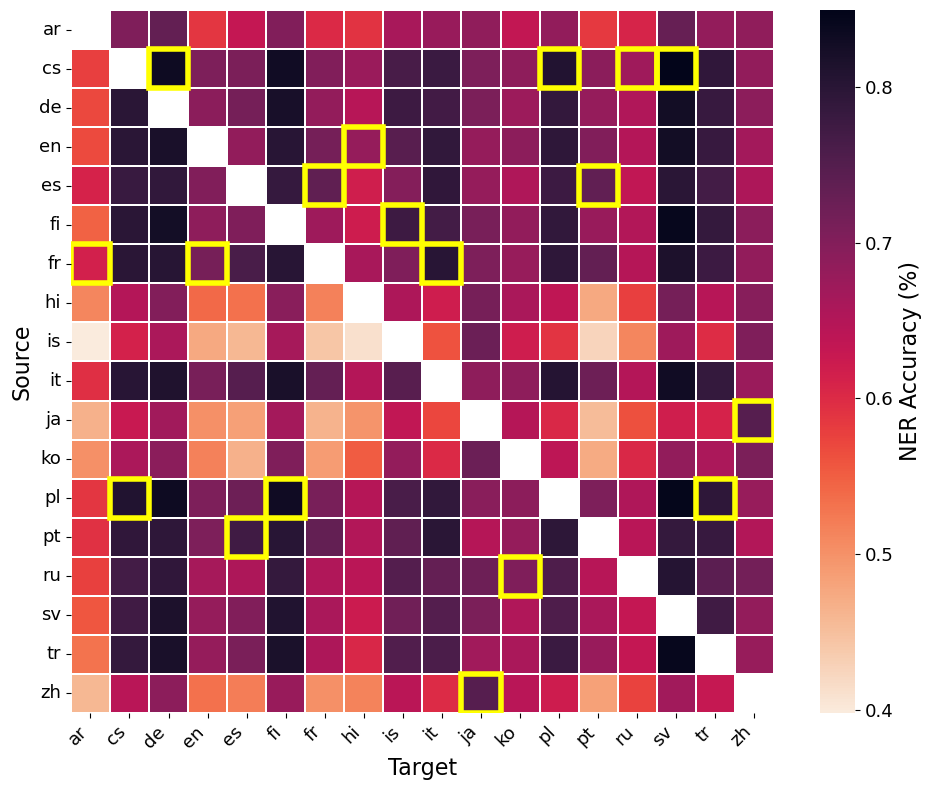}
  \includegraphics[width=0.48\linewidth]{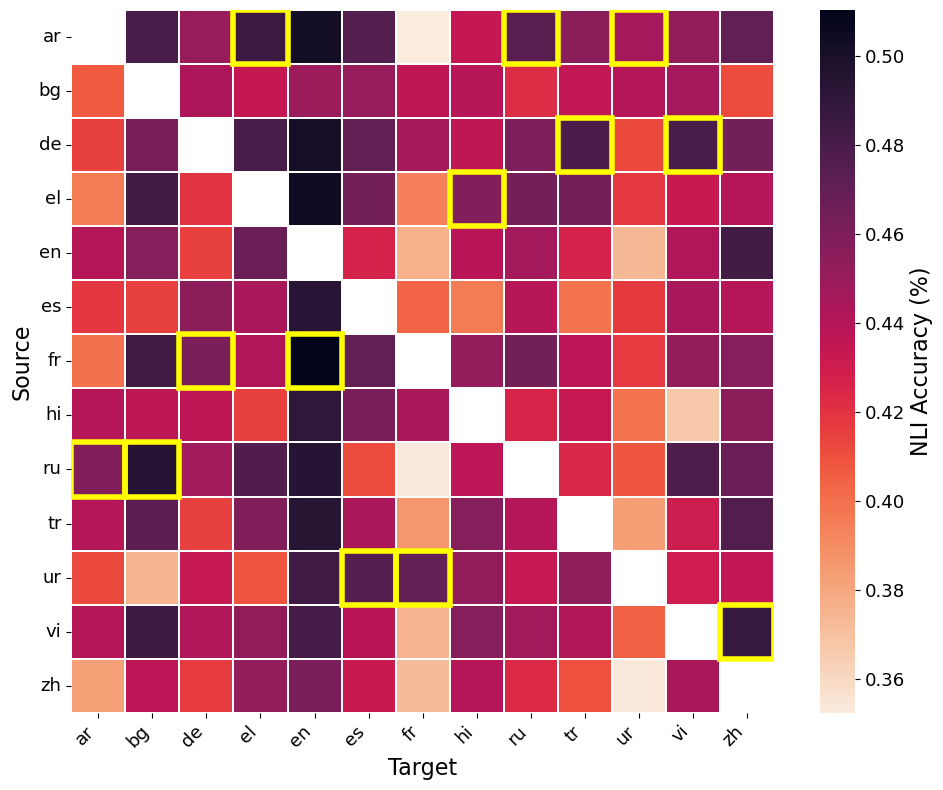}
  \caption{Cross-lingual probing accuracy at the final checkpoint for POS (top left), NER (top right), and XNLI (bottom). Each cell shows accuracy of a probe trained on the source language (row) and evaluated on the target (column). Best source for each target is highlighted in yellow.}
  \label{fig:crosslingual_matrix}
\end{figure*}

Figure~\ref{fig:crosslingual_matrix} shows the full cross-lingual probing results for POS, NER, and XNLI at the final checkpoint. Each heatmap shows transfer accuracy from a source language (row) to a target language (column). The highest-performing source language for each target is highlighted in yellow.

\section{Visualization of Language Discrimination Effects on Cross-Lingual Performance}\label{appendix:LD_correl_pos_ner}

\begin{figure*}[h]
\centering
  \includegraphics[width=0.48\linewidth]{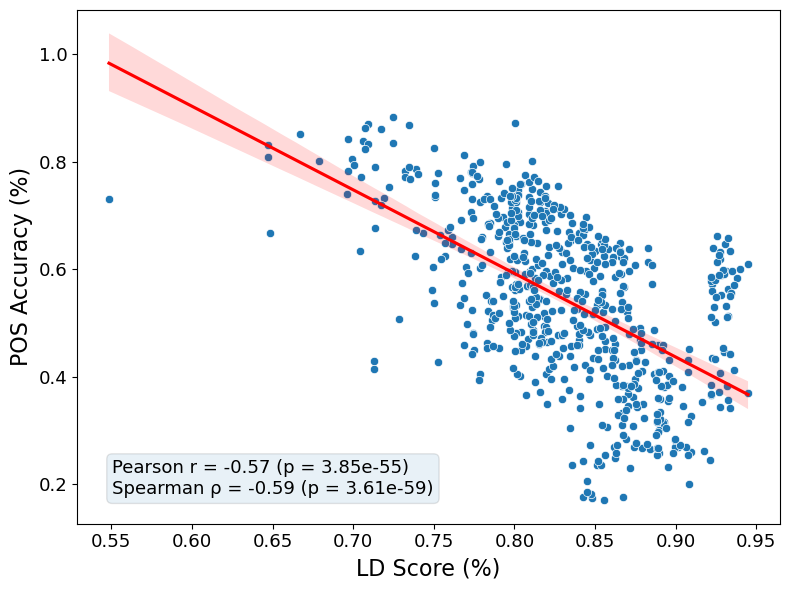} \hfill
  \includegraphics[width=0.48\linewidth]{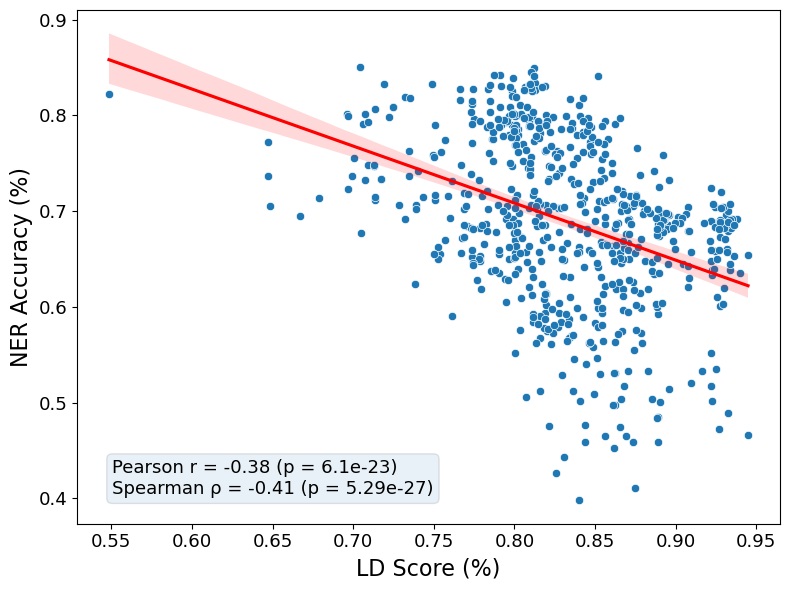}
\caption{Relationship between language discrimination scores and cross-lingual transfer accuracy for POS tagging (left) and NER (right) across all source-target language pairs. Each point represents a language pair, with the x-axis showing the language discrimination score and the y-axis showing transfer accuracy. The downward trend demonstrates that higher language discrimination (more distinct language forms) is associated with lower cross-lingual transfer performance for POS and NER.}
\label{fig:ld_transfer_corr}
\end{figure*}

Figure \ref{fig:ld_transfer_corr} illustrates the relationship between language discrimination scores and cross-lingual transfer accuracy for all source-target language pairs in our experiments. For both POS tagging and NER tasks, we observe a strong negative correlation: language pairs with higher discrimination scores (indicating more distinct linguistic forms) consistently show lower transfer performance. This visualization reinforces our regression findings that language discrimination acts as a significant negative predictor of cross-lingual transfer success.

The scatter plots reveal that when models encode languages in ways that make their forms highly distinguishable from each other, their ability to transfer knowledge between those languages for form-focused task as POS and NER diminishes. Conversely, when language forms are less discriminable (more shared or mixed representations), cross-lingual transfer improves. 

\section{ABX-Guided Source Language Selection}
\label{appendix:abx-source-selection}

Inspired by our earlier use of ABX scores to guide checkpoint selection (Section~\ref{appendix:abx-checkpoint-selection}), we investigate whether ABX language discrimination can also inform source language selection in cross-lingual transfer. Specifically, for each target language, we test whether the source language with the lowest ABX language discrimination score yields the highest transfer performance.

\paragraph{Exact Match and Top-k Accuracy}  
We first compare, for each target language, the true best source (i.e., the one yielding the highest transfer accuracy) with the ABX-selected source (i.e., the one with minimal ABX LD). Exact matches occur in 2/18 (POS) and 7/18 (NER) cases. When considering the top-3 sources, ABX guidance succeeds in 6/18 (POS) and 12/18 (NER) cases, suggesting it often identifies competitive transfer candidates.

\paragraph{Comparison to Random Selection}  
Next, we evaluate how ABX-guided selection compares to a naive random baseline. For each target, we compare the transfer accuracy of the ABX-selected source to that of 100 randomly sampled sources, and compute the proportion of wins. The ABX-guided source outperforms a random one in 73.0\% ± 27.5\% of trials for POS, and 84.8\% ± 23.4\% for NER.

\begin{figure*}[h]
    \centering
    \includegraphics[width=0.8\linewidth]{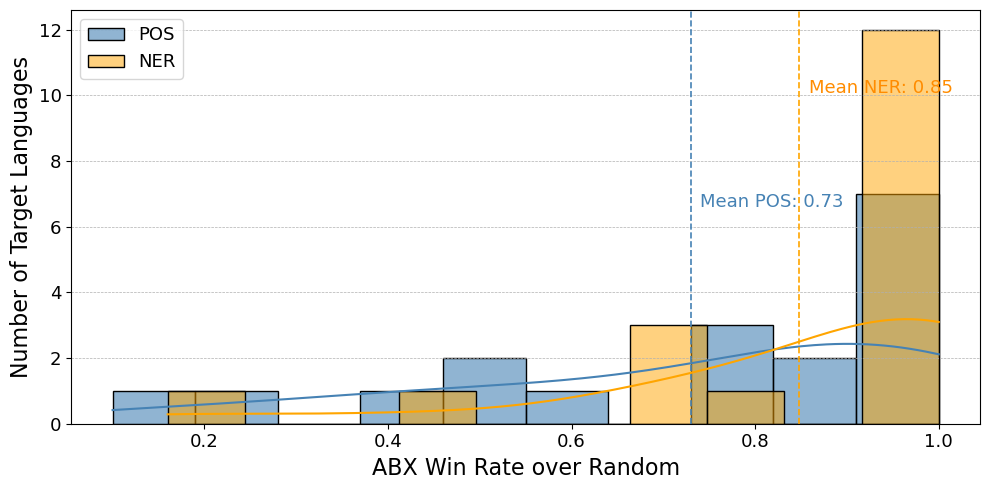}
    \caption{Distribution of ABX win rates across target languages for POS (blue) and NER (orange).
Dashed lines indicate average win rate per task. A value above 0.5 reflects better-than-random performance.}
    \label{fig:abx_winrate_hist}
\end{figure*}

Figure~\ref{fig:abx_winrate_hist} shows the full distribution of these per-target win rates. Most values exceed 70–80\%, and very few fall below the 50\% chance level, indicating that ABX LD offers a consistent and effective heuristic for source selection.

\paragraph{Conclusion}  
While ABX-guided source selection does not always identify the single best transfer source, it reliably outperforms random baselines. Compared to typological or lexical similarity heuristics (which are often noisy or task-specific) ABX LD offers a simple, data-driven alternative for identifying effective source languages in cross-lingual transfer.

